\documentclass[10pt,twocolumn,letterpaper]{article}

\usepackage[pagenumber]{cvpr} 

\usepackage[dvipsnames]{xcolor}

\definecolor{cvprblue}{rgb}{0.21,0.49,0.74}
\usepackage[pagebackref,breaklinks,colorlinks,citecolor=cvprblue]{hyperref}

\def\paperID{6283} 
\def\confName{CVPR}
\def\confYear{2024}

\usepackage{colortbl}
\usepackage{multirow}
\usepackage[linesnumbered, ruled]{algorithm2e}
\usepackage{makecell}
\usepackage[accsupp]{axessibility}

\title{NetTrack: Tracking Highly Dynamic Objects with a Net}

\author{Guangze Zheng$^{1}$ \quad Shijie Lin$^{1}$ \quad  Haobo Zuo$^{1}$  \quad  Changhong Fu $^{2}$  \quad  Jia Pan$^{1}$\footnotemark \\
$^{1}$The University of Hong Kong \quad $^{2}$Tongji University  \\
{\tt\small \{guangze, lsj2048, haobozuo\}@connect.hku.hk, changhongfu@tongji.edu.cn, jpan@cs.hku.hk}\\
\href{https://george-zhuang.github.io/nettrack}{ \tt\small https://george-zhuang.github.io/nettrack}
}

\begin{document}

\twocolumn[{
\renewcommand\twocolumn[1][]{#1}
\maketitle
\begin{center}
\centering
\vspace{-10pt}
\includegraphics[width=\textwidth]{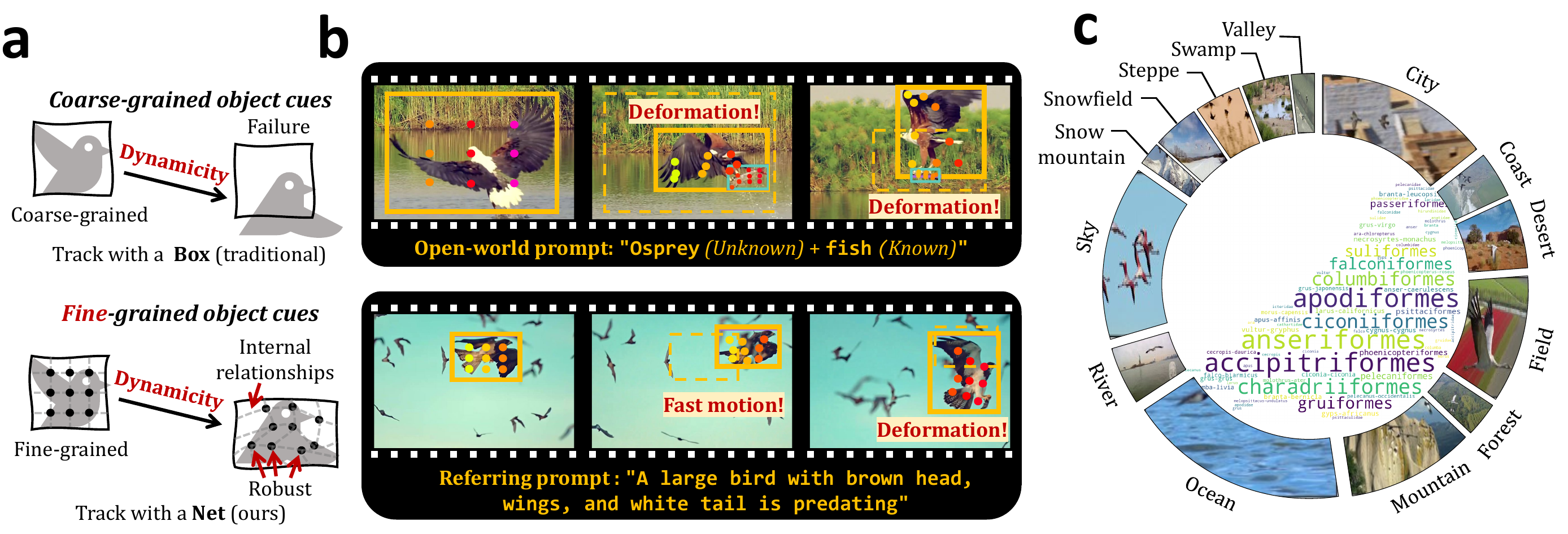}
\captionof{figure}{
    \textbf{a} The visualization of the proposed NetTrack is similar to a \textbf{Net}. Object dynamicity distorts the internal relationships of the object, presenting challenges for traditional coarse-grained tracking methods that rely solely on bounding boxes. While NetTrack introduces fine-grained Nets that are robust to dynamicity.
    \textbf{b} Qualitative results of NetTrack tracking highly dynamic objects under \textit{open-world tracking} and \textit{referring expression comprehension} settings. Dynamicity like deformation and fast motion results in drastic changes in the coarse-grained representation, while the fine-grained Nets can contract robustly. 
    The dashed boxes represent the object position from the previous time step.
    \textbf{c} We propose a challenging benchmark named BFT, dedicated to evaluating highly dynamic object tracking with abundant scenarios shown in the external circular and diverse species shown in the central word cloud.}
\label{fig:1}
\end{center}
}]\footnotetext{$^*$ Corresponding author.}

\begin{abstract}
The complex dynamicity of open-world objects presents non-negligible challenges for multi-object tracking (MOT), often manifested as severe deformations, fast motion, and occlusions.
Most methods that solely depend on coarse-grained object cues, such as boxes and the overall appearance of the object, are susceptible to degradation due to distorted internal relationships of dynamic objects.
To address this problem, this work proposes \textbf{NetTrack}, an efficient, generic, and affordable tracking framework to introduce fine-grained learning that is robust to dynamicity. Specifically, NetTrack constructs a dynamicity-aware association with a fine-grained \textbf{Net}, leveraging point-level visual cues. Correspondingly, a fine-grained sampler and matching method have been incorporated. Furthermore, NetTrack learns object-text correspondence for fine-grained localization.
To evaluate MOT in extremely dynamic open-world scenarios, a bird flock tracking (\textbf{BFT}) dataset is constructed, which exhibits high dynamicity with diverse species and open-world scenarios. Comprehensive evaluation on BFT validates the effectiveness of fine-grained learning on object dynamicity, and thorough transfer experiments on challenging open-world benchmarks, i.e., TAO, TAO-OW, AnimalTrack, and GMOT-40, validate the strong generalization ability of NetTrack even without finetuning.

\end{abstract}    
\section{Introduction}
\label{sec:intro}

Multiple object tracking (MOT)~\cite{ovtrack,ocsort,video-owlvit,tao-ow,luo2021multiple} aims to maintain continuous visual perception of objects of interest in videos and the real world.
Traditional MOT methods often assume objects as coarse-grained entities because in classical MOT tasks~\cite{dendorfer2020mot20,yu2020bdd100k,waymo}, the dynamicity of specific object categories~\cite{dendorfer2021motchallenge} and scenes is not significant, and the internal relationships within objects are relatively stable.
However, the demand for tracking arbitrary objects, especially highly dynamic objects, in open-world MOT tasks~\cite{ling2019behavioural,masmitja2023dynamic,lee2021visual} severely challenges this assumption.

The high dynamicity of open-world objects, manifested as severe deformation, fast motion, and frequent occlusion, poses challenges for existing methods in two major aspects:
\begin{enumerate}
    \item[1)] 
    \textbf{Association}~~For most methods relying solely on coarse-grained visual representations, the high dynamicity renders the temporal continuity fragile in terms of association, since the internal relationships in the objects are distorted. These methods typically represent the overall object as coarse-grained bounding boxes~\cite{bytetrack,ocsort} or the corresponding features~\cite{ovtrack,video-owlvit}, and the dynamicity significantly reduces the similarity of such representations across different time steps, as shown in \cref{fig:1}-b.
    \item[2)] 
    \textbf{Localization}~~The high dynamicity also poses challenges to establishing accurate object-text correspondence for localization. State-of-the-art (SoTA) methods~\cite{ovtrack,video-owlvit} typically learn the coarse-grained correspondence between the entire image and text in pre-training.
    For severely deformed or occluded objects, these methods often struggle to localize.
\end{enumerate}

In this work, we propose NetTrack, introducing fine-grained learning to address the above two aspects.
Regarding association, NetTrack utilizes physical points on the object's appearance that are less susceptible to object dynamicity and form fine-grained visual cues. For localization, grounded pre-training is utilized to learn fine-grained correspondences between objects and text. Therefore, our primary contributions are outlined as follows:

\noindent\textbf{{Fine-grained Net for dynamicity-aware association}}
~~
Instead of viewing the object as a coarse-grained entity, this work tracks the object with a fine-grained Net, which leverages points of interest (POIs) on the surface of object appearance. The dynamicity, such as deformations, distorted internal relationships between POIs by altering global relative position and appearance feature distribution, while the fine-grained representations of the points themselves, such as local appearance color and relationships with neighboring points, are seldom affected and exhibit robustness, as shown in \cref{fig:1}-b. 
Following this viewpoint, we design a fine-grained sampler to discover potential POIs and utilize fine-grained visual cues of these points, along with the emerging physical point tracking methods~\cite{cotracker,tapir,pips}, for robust tracking. Subsequently, a simple yet effective fine-grained similarity calculation method is proposed to determine the containment relationship between the tracked POIs and candidate objects. The proposed fine-grained similarity scores are combined with the existing coarse-grained to achieve more robust association of dynamic objects.

\begin{figure}[t!]
	\begin{center}
		\includegraphics[width=0.475\textwidth]{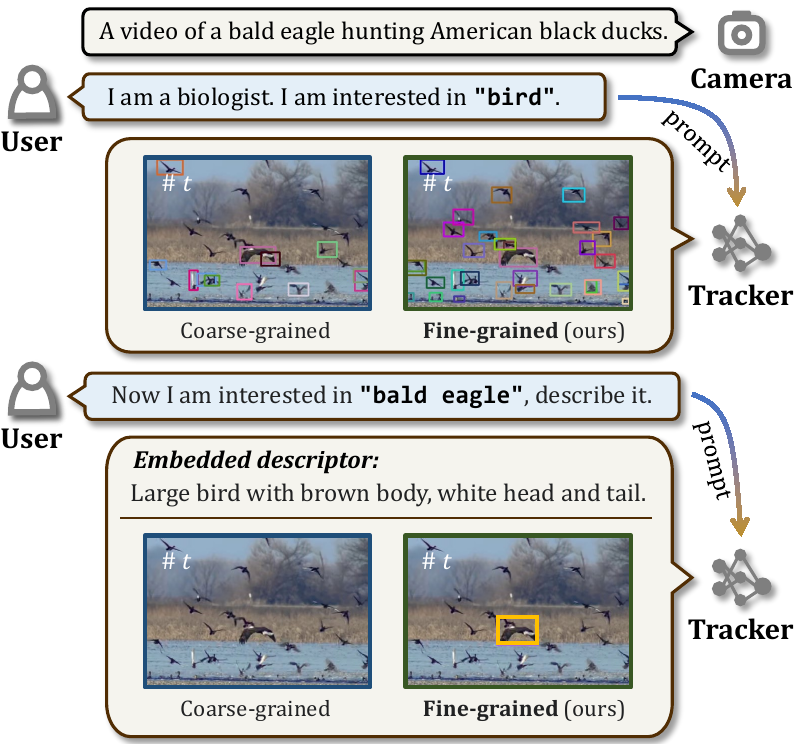}
	\end{center}
        \vspace{-10pt}
	\caption{Comparison between the localization method~\cite{owl-vit,video-owlvit} based on coarse-grained object-text correspondence and our fine-grained method. Our fine-grained approach localizes dynamic objects better and can leverage professional descriptions from embedded descriptors (GPT-3.5~\cite{gpt} in the example) with a better understanding of context.}
	\label{fig:det}
        \vspace{-10pt}
\end{figure}

\noindent\textbf{{Object-text correspondence for fine-grained localization}} 
~~
To discover and localize highly dynamic objects of interest in tracking, this work adopts a pre-training method to tracking by phrase grounding~\cite{glip,glipv2,groundingdino} for fine-grained object-text correspondence. Compared to CLIP-based tracking methods~\cite{video-owlvit,ovtrack} that utilize coarse-grained image-text correspondence, NetTrack can more effectively distinguish highly dynamic objects, as shown in \cref{fig:det}. Furthermore, by embedding descriptors (GPT-3.5~\cite{gpt} in \cref{fig:det}) within the framework, the proposed framework learns contextual information, such as provided professional application and knowledge context, to mitigate background interference and achieve practical real-world applications for efficient dynamic object tracking.

\noindent\textbf{{A highly dynamic benchmark and transfer experiments on diverse scenarios}}
~~ 
This work introduces a highly dynamic open-world MOT dataset, named bird flock tracking (BFT), to evaluate the performance of tracking methods in tracking highly dynamic objects. BFT is particularly notable for the complex and unpredictable dynamicity of 22 bird species for three main reasons: 1) Fast motion caused by the three-dimensional activity space. 2) Deformations resulting from frequent flapping of wings~\cite{ling2018simultaneous}. 3) Occlusions arising from the collective behavior of birds~\cite{ling2019behavioural, ling2019costs} in the flock. Furthermore, BFT comprises 14 distinct open-world scenes and 22 species in 106 sequences, showcasing a rich diversity, as depicted in \cref{fig:1}-c. 
In evaluation, the proposed NetTrack framework reaches SoTA performance in tracking highly dynamic objects in BFT. Besides, comprehensive zero-shot transfer experiments show NetTrack surpasses tracking baselines on several challenging open-world MOT benchmarks, \eg, TAO~\cite{dave2020tao,ovtrack}, TAO-OW~\cite{tao-ow}, AnimalTrack~\cite{zhang2022animaltrack}, and GMOT-40~\cite{bai2021gmot}. The introduced fine-grained learning contributes to stronger generalization ability of NetTrack even without finetuning.
As an efficient, generic, and affordable tracking framework, NetTrack also exhibits potential in open-world application scenarios, further highlighting its suitability for downstream tasks.

\section{Related Work}
\label{sec:related_work}
\noindent\textbf{Open-world multi-object tracking methods}
~~
Tracking-by-detection~\cite{sort, ioutracker, bytetrack, ocsort,deepsort,trackformer} has been the most popular framework in MOT, which includes localizing potential objects and associating them over time~\cite{luo2021multiple}. Traditional MOT methods typically focus on limited scenes and object categories, such as pedestrians~\cite{sort,deepsort,fairmot,transtrack} in public places or vehicles~\cite{bytetrack,qdtrack} in autonomous driving scenarios. Comparatively, open-world tracking tasks require trackers to have the ability to track any object in complex and dynamic scenes. The rise of CLIP~\cite{clip}-based open-set object detection~\cite{vild,regionclip,owl-vit} has promoted this task, inspiring advanced open-world tracking baselines~\cite{ovtrack,owl-vit} to utilize CLIP-style pre-training to achieve generalization by leveraging the correspondence between text and images. However, these mainstream tracking methods~\cite{transtrack,bytetrack,sort,ocsort,ovtrack} usually view objects as coarse-grained bounding boxes, but the high dynamicity of open-world objects can often disrupt the temporal similarity of this coarse representation. 
Moreover, compared to the shallow-fused vision-language features used in CLIP-like pre-training, localizing dynamic objects often requires establishing fine-grained correspondences between the object and text to counteract the appearance distortion or impairment of the objects.

The recent emergence of physical point tracking methods~\cite{pips, tapvid, omnimotion, tapir, cotracker} has inspired this work to introduce fine-grained visual cues of objects. These methods aim to track arbitrary physical points over video clips, relying on point-level appearance representation rather than coarsely propagating the overall object, therefore holding promise to maintain good generalization for dynamic objects. Additionally, the pre-training approach based on phrase grounding~\cite{phrasegrounding} has also been applied in open-set object detection tasks~\cite{glip,glipv2,groundingdino}, and its potential benefits for dynamic object tracking are anticipated due to object-level, language-aware, and semantic-rich visual representations.

\noindent\textbf{Open-world multi-object tracking benchmarks}
~Classical MOT benchmarks mainly focus on limited object categories and scenarios, where objects typically maintain stable appearances or poses and undergo relatively simple motion, \eg, tracking
pedestrian~\cite{leal2015motchallenge,milan2016mot16,dendorfer2020mot20,dendorfer2021motchallenge} or vehicles~\cite{geiger2012we,yu2020bdd100k,sun2020scalability}. With increasing demands for open-world tracking applications, MOT benchmarks that focus on a wider range of scenarios and object classes have emerged. TAO~\cite{dave2020tao} includes numerous \textit{unseen} objects in massive data, GMOT-40~\cite{bai2021gmot} focuses on tracking \textit{unseen} object categories, and AnimalTrack~\cite{zhang2022animaltrack} places emphasis on tracking wildlife. Later, TAO-OW~\cite{tao-ow} defines \textit{known} and \textit{unknown} object categories in an open-world setting, and Li \etal~\cite{ovtrack} divide object categories into \textit{base} and \textit{novel} on the TAO benchmark in an open-vocabulary setting. In diverse open-world MOT tasks, while learning \textit{unseen} classes is of paramount importance, the dynamicity resulting from potential severe deformations and fast motion of these \textit{unseen} objects are equally crucial, necessitating a comprehensive evaluation.
\section{Method}
The proposed NetTrack framework introduces a fine-grained Net for dynamicity-aware object association and fine-grained object-text correspondence for dynamicity-aware localization.
\cref{sec:asso} describes structuring the object into fine-grained Nets using sampling and performing association.
\cref{sec:det} primarily discusses how fine-grained object-text correspondence positively affects the localization of dynamic objects.

\subsection{Fine-Grained Net}\label{sec:asso}
\begin{figure}[t!]
	\begin{center}
		\includegraphics[width=0.47\textwidth]{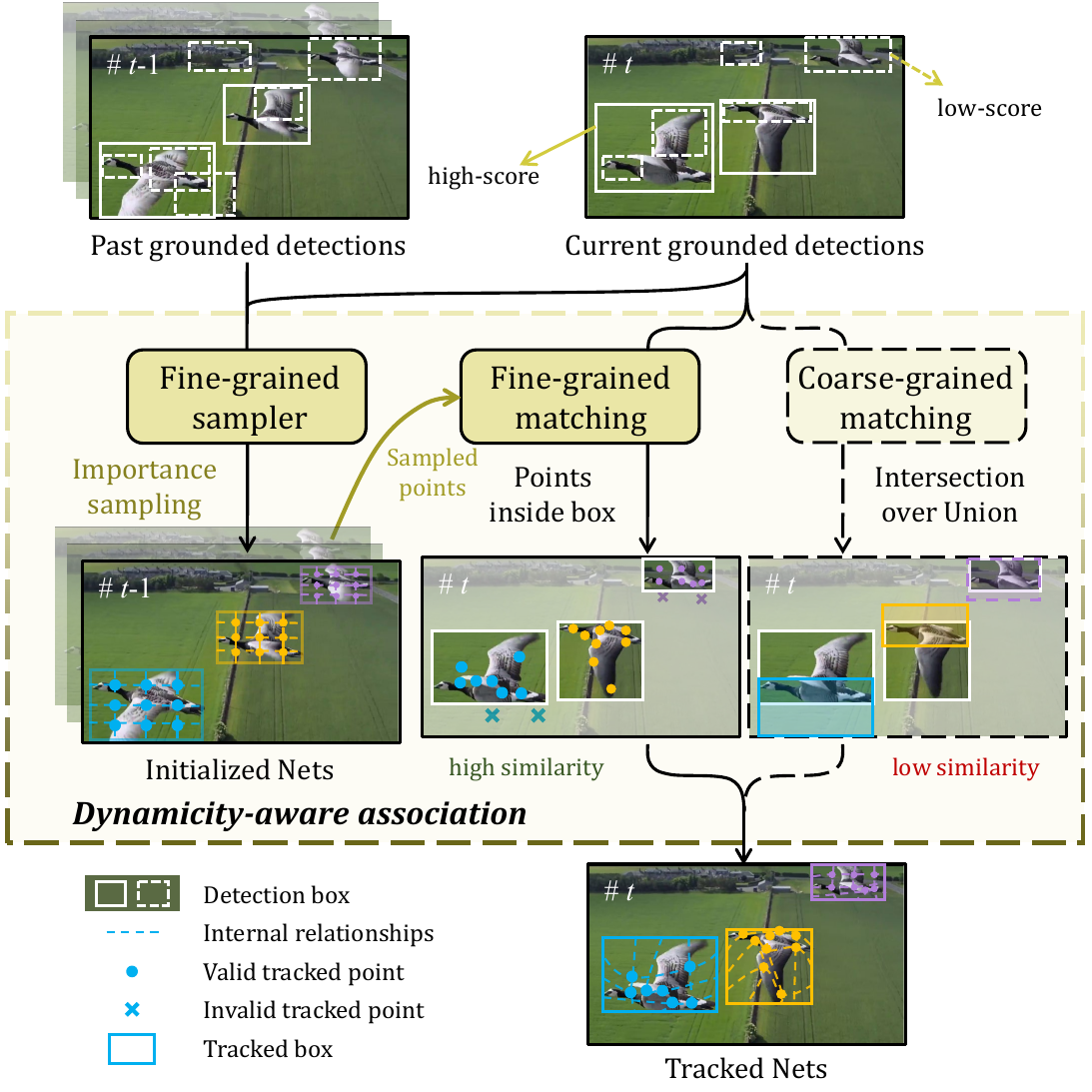}
	\end{center}
	\vspace{-10pt}
	\caption{Dynamicity-aware association in the NetTrack framework. Unlike the coarse-grained association methods that only learn the box motion or overall appearance, dynamicity-aware association benefits from fine-grained Nets which are robust against the open-world dynamicity and exhibit stronger generalization ability.}
	\label{fig:asso}
	\vspace{-10pt}
\end{figure}

The proposed dynamicity-aware association utilizes fine-grained Nets to construct robust visual cues for object dynamicity. It mainly consists of a fine-grained sampler and a matching method. The overall process is shown in \cref{fig:asso}.

\noindent\textbf{Fine-grained sampler} 
~~
This work introduces point-level visual cues to form fine-grained Nets with points of interest (POIs).
Ideally, sampling POIs should accurately capture every valuable point on the surfaces of every interested object, avoiding background interference or the redundant computational burden. A straightforward thought is to sample POIs inside boxes of tracked objects and update points frame by frame. However, such an approach can impose a certain computational burden, ignorance of false negative samples, and insufficient visual context. Therefore, a fine-grained sampler is proposed to sample cross-frame POIs.

Denote the expected distribution of POIs as $f(\mathbf{x})$, where $\mathbf{x}$ refers to a point in the image $I$. The object motion is estimated based on Kalman Filter~\cite{Kalmanfilter} as in~\cite{sort,ioutracker,bytetrack,ocsort}. Such estimation acts as the coarse distribution of novel objects in a certain period of $\mathcal{S}$ frames. The distribution can then be transformed to a point-level form as $p(\mathbf{x}|\mathcal{T}^\mathrm{coarse}_\mathrm{o}, \{\mathbf{I}\}_{i=1}^{\mathcal{S}})$, where $\mathcal{T}^\mathrm{coarse}_\mathrm{o}$ is the coarsely estimated coarse-grained trajectory of objects and $p(\cdot)$ is a binary distribution that discover the potential POIs. This distribution serves as an importance weight to sample the POIs. Given point number $K$, the expected POIs can then be formulated using importance sampling~\cite{importance} as:
\small
\begin{equation}
	\mathbb{E}_{\mathbf{x} \sim p(\mathbf{x|\{\mathbf{I}\}_{i=1}^{\mathcal{S}}})}[f(\mathbf{x})] = \frac{1}{K}\sum^{K}_{i=1}\frac{f(\mathbf{x}_i)}{p(\mathbf{x}_i|\mathcal{T}^\mathrm{coarse}_\mathrm{o},\{\mathbf{I}\}_{i=1}^{\mathcal{S}})} ~.
\end{equation}
\normalsize

Therefore, the fine-grained POIs are determined at frame \#$t$-1 and estimated at frame \#$t$ with point tracking models.

\noindent\textbf{Fine-grained matching} 
~~
Utilizing fine-grained Nets for tracking requires matching the memorized POIs with the current detection results based on temporal similarity. Given the point tracker model \texttt{Tr}$_\mathrm{p}$, the estimated point trajectories $\mathcal{T}_\mathrm{p}$ can be obtained in the aforementioned period. After acquiring the detection results $\mathcal{D}_t$ for the current frame $\#t$, the fine-grained matching method calculates the number of estimated points from a Net that fall inside a candidate detection box as fine-grained similarity. Suppose $N$ is the number of tracked objects in frame $\#t-1$, the element $\mathbf{S}_{i,j}$ of the matching fine-grained score matrix $\mathbf{S}$ of $N$ Nets $\{\mathbf{P}_i\}_{i=1}^N$ and $M$ detection boxes $\{\mathbf{b}_j\}_{j=1}^M$ can be expressed as:
\small
\begin{equation}
	\mathbf{S}_{i,j} =  w_{i,j}\frac{ |\mathbf{P}_i\cap \mathbf{b}_j|}{|\mathbf{P}_i|}~~, \quad
	w_{i,j}=\min\{1, \frac{\mathcal{A}(\hat{\mathbf{b}}_i)}{\mathcal{A}(\mathbf{b}_j)}\}~~,
\end{equation}
\normalsize
where $w$ is a weight to penalize candidate detection boxes with excessively large areas, as larger areas often result in predicted points prone to fall within the box, leading to potential misjudgments. $|\mathbf{P}_i\cap \mathbf{b}_j|$ refers to the number of points from Net $\mathbf{P}_i$ positioning inside $\mathbf{b}_j$, depicted as valid points in~\cref{fig:asso}, and $|\mathbf{P}_i|$ is the number of points in Net $\mathbf{P}_i$. $\mathcal{A}(\cdot)$ refers to area of a box, and $\hat{\mathbf{b}}$ is the predicted box of a tracked object in frame $\#t$ using~\cite{Kalmanfilter}. 
Afterward, combined with the coarse-grained similarity score, the overall matching score can be obtained. As shown in \cref{fig:asso}, object dynamicity often leads to a decrease in coarse-grained similarity in Intersection over Union (IOU), while fine-grained association remains robust.
The matching process is then carried out using the Hungarian algorithm~\cite{kuhn1955hungarian}. Details of the method are described in~\cref{algorithm}.

\subsection{Fine-Grained Object-Text Correspondence}\label{sec:det}
To learn fine-grained object-text correspondence for localization, this work introduces a pre-training strategy based on phrase grounding to track dynamic objects and mitigates the adverse effects of object dynamics with a deep fusion of textual and object features.
Different from SoTA tracking methods~\cite{video-owlvit,ovtrack} that utilizes CLIP~\cite{clip}-based pre-training, we follow~\cite{glip,glipv2,groundingdino} to identify the correspondence between phrases in sentences and objects in images to formulate fine-grained object-text correspondence. Given the input image $\mathbf{I}$ and language prompt $\mathbf{P}$, corresponding object features $\mathbf{F}_{\text{O}}$ and language features $\mathbf{F}_{\text{L}}$ can be obtained with a visual encoder $\text{Enc}_{\text{V}}$ and a language encoder $\text{Enc}_{\text{L}}$, respectively. Afterward, we can get fused features $\mathbf{F}'_{\text{O}}$ and $\mathbf{F}'_{\text{L}}$ by deep fusion, and further obtain the object-text correspondence score $\mathbf{S}_\text{ground}$. The formula for this process is:

\begin{equation}
	\begin{split}
	&\mathbf{F}_{\text{O}} = \text{Enc}_{\text{V}}(\mathbf{I}), ~~\mathbf{F}_{\text{L}}=\text{Enc}_{\text{L}}(\mathbf{P}), \\
	&\mathbf{F}'_{\text{O}},~\mathbf{F'}_{\text{L}} = \text{Fuse}(\mathbf{F}_{\text{O}}, \mathbf{F}_{\text{L}}),~~\mathbf{S}_\text{ground}=\mathbf{F}'_{\text{O}}\mathbf{F}'^\top_{\text{L}}.
	\end{split}
\end{equation}

From a visual perspective, fine-grained object-text correspondence enhances the language awareness of visual features, thereby enabling better discernment of deformed objects. From a language view, such correspondence learns contextualized representations at the word or sub-sentence level during pre-training~\cite{groundingdino}, avoiding biases caused by unnecessary word interactions. The proposed framework also allows for a more detailed understanding of the object with an embedded descriptor, \eg, large language models~\cite{gpt,gpt4}. Consequently, such fine-grained correspondence is better suited for capturing more specific contextual information in professional scenarios, as illustrated in \cref{fig:det} and \cref{fig:qualitative}.

\begin{figure*}[t!]
	\begin{center}
		\includegraphics[width=0.95\textwidth]{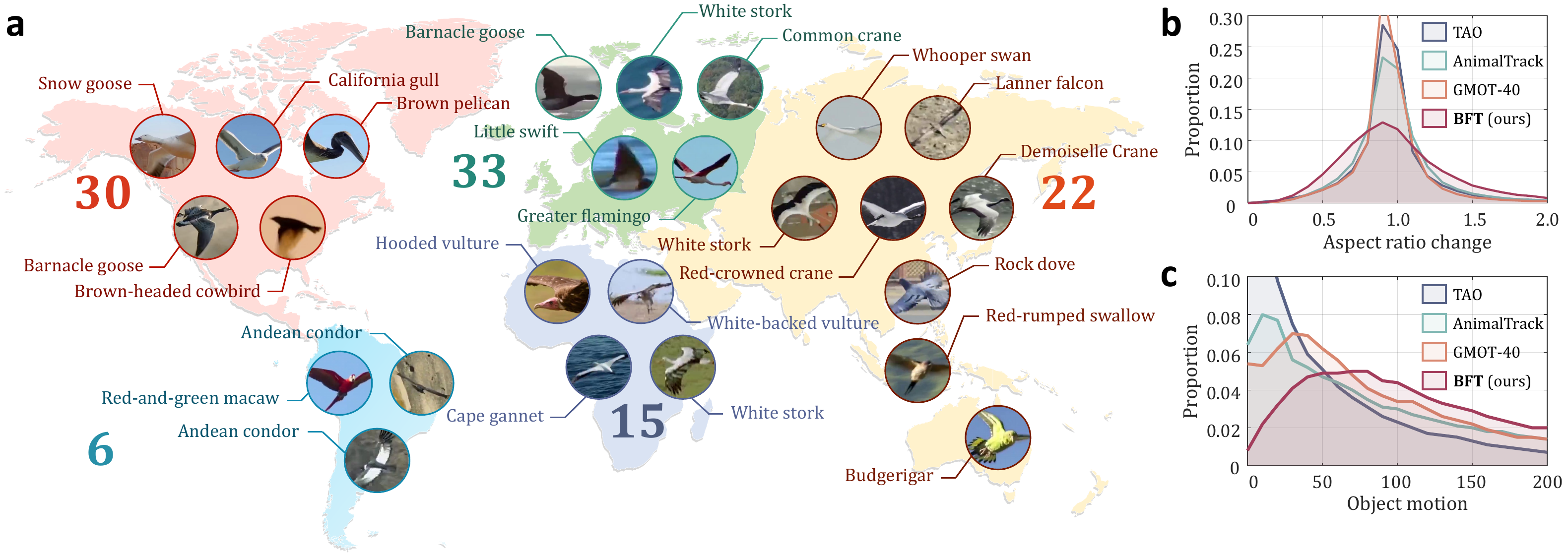}
	\end{center}
         \vspace{-10pt}
	\caption{\textbf{a} The diverse geographical distribution of some representative flying bird species exhibits the diversity of BFT. The numbers on the map represent the number of videos in each corresponding area, \eg, 30 videos from North America. \textbf{b} 
 Dynamicity comparison between BFT and other datasets on aspect ratio change. The more dispersed distribution means more frequent object deformation and occlusion in BFT. \textbf{c} Dynamicity comparison between BFT and other datasets on object motion. The larger object motion in BFT represents the faster motion of objects.
    }
	\label{fig:dataset}
        \vspace{-10pt}
\end{figure*}

\section{BFT Dataset}\label{sec:bft_dataset}

\noindent{\textbf{Data collection}}~
Bird flocks are among the most dynamic objects to track in the open world and thus are considered ideal subjects for this work. The dynamicity of birds is mainly attributed to three phenomena: 1) Bird flocks exhibit higher maneuverability compared to ground objects due to the three-dimensional activity space and an additional degree of freedom. In addition, the inertia of birds is relatively small, allowing them to accelerate, decelerate, and change direction more flexibly. The complex aerodynamic effects~\cite{fry2003aerodynamics} also make the motion of flying bird flocks more difficult to predict. 2) Birds generally experience frequent and intense deformation during flight, mainly due to wingbeat~\cite{ling2018simultaneous}. 3) Collective behavior~\cite{ling2019behavioural, ling2019costs} is widespread in many species of bird flocks. This often results in a dense distribution of bird flocks within a limited space, making it visually susceptible to occlusion. In addition to the aforementioned dynamic challenges, birds often have similar appearances in flocks, which also adds to the difficulty of visual discrimination.

To showcase the diversity of open-world scenarios and the variety of species, the BFT dataset incorporates 22 bird species and 14 common natural and cultural scenes, covering six continents as illustrated in \cref{fig:dataset}-a and \cref{fig:1}-c. Detailed corresponding order, family, genus, and species of bird are in \cref{fig:order}. 
The primary data source is the BBC nature documentary series \textit{Earthflight}~\cite{earthflight}. 106 carefully selected clips were extracted from around 6 hours of video, which were further divided into a training set with 35 videos, a validation set with 25 videos, and a test set with 36 videos. All the data underwent meticulous annotation by experts and multiple rounds of review by tracking domain experts, as well as verification by experts in the field of biology. The frame rate of both the videos and annotations is commonly set at 25 frames per second (FPS).

\noindent\textbf{High dynamicity}~
The higher dynamicity in BFT includes more severe deformations, faster motion, and more frequent occlusions. Quantitatively, \cref{fig:dataset}-b,c compare the dynamicity of BFT with other open-world MOT datasets~\cite{dave2020tao,bai2021gmot,zhang2022animaltrack} from two aspects. Specifically, aspect ratio change (ARC)~\cite{trackingnet, lasot} is a commonly used tracking attribute, which measures the frequency and severity of object deformations or occlusions. Object motion is another attribute to measure the displacement of an object between two consecutive time steps. Detailed statistics are shown in \cref{sec:benchmak_details}.
Due to the more dispersed ARC distribution of BFT and the larger values of the motion distribution, BFT represents stronger dynamicity compared to other dataset.

\section{Experiments}
\begin{table*}[t]
	\centering
	\caption{Overall evaluation on the highly dynamic BFT. 
		$^\star$ denotes comprehensive evaluation metrics. Finetuned results are in \textcolor[rgb]{0.5, 0.5, 0.5}{gray}, and the best results in each setting are in \textbf{bold}. $^\dagger$ denotes the offline setting.}
	\resizebox{0.85\linewidth}{!}{
		\begin{tabular}{ccccccccccccc}
			\toprule
			\multirow{2}[4]{*}{\textbf{Detector}} & \multirow{2}[4]{*}{\textbf{Method}} & \multicolumn{10}{c}{\textbf{BFT benchmark  evaluation}}                       & \multirow{2}[4]{*}{\textbf{Line}} \\
			\cmidrule{3-12}          &       & OWTA$^\star$↑ & D. Re.↑ & D. Acc.↑ & D. Pr.↑ & A. Acc.↑ & A. Re.↑ & A. Pr.↑ & HOTA$^\star$↑ & MOTA$^\star$↑ & IDF1$^\star$↑  &  \\
			\midrule
			\multicolumn{13}{l}{\textcolor[rgb]{ .5,  .5,  .5}{\textit{a) Finetuned on BFT dataset}}} \\
			\midrule
			\textcolor[rgb]{ .5,  .5,  .5}{YOLO-v3~\cite{yolov3}} & \textcolor[rgb]{ .5,  .5,  .5}{JDE~\cite{jde}} & \textcolor[rgb]{ .5,  .5,  .5}{33.0} & \textcolor[rgb]{ .5,  .5,  .5}{46.9} & \textcolor[rgb]{ .5,  .5,  .5}{40.9} & \textcolor[rgb]{ .5,  .5,  .5}{61.0} & \textcolor[rgb]{ .5,  .5,  .5}{23.4} & \textcolor[rgb]{ .5,  .5,  .5}{24.7} & \textcolor[rgb]{ .5,  .5,  .5}{66.7} & \textcolor[rgb]{ .5,  .5,  .5}{30.7} & \textcolor[rgb]{ .5,  .5,  .5}{35.4} & \textcolor[rgb]{ .5,  .5,  .5}{37.4} & \textcolor[rgb]{ .5,  .5,  .5}{1} \\
			\midrule
			\multirow{2}[2]{*}{\textcolor[rgb]{ .5,  .5,  .5}{CenterNet~\cite{centernet}}} & \textcolor[rgb]{ .5,  .5,  .5}{CenterTrack~\cite{centertrack}} & \textcolor[rgb]{ .5,  .5,  .5}{61.6} & \textcolor[rgb]{ .5,  .5,  .5}{70.5} & \textcolor[rgb]{ .5,  .5,  .5}{58.5} & \textcolor[rgb]{ .5,  .5,  .5}{71.5} & \textcolor[rgb]{ .5,  .5,  .5}{54.0} & \textcolor[rgb]{ .5,  .5,  .5}{57.8} & \textcolor[rgb]{ .5,  .5,  .5}{82.8} & \textcolor[rgb]{ .5,  .5,  .5}{56.0} & \textcolor[rgb]{ .5,  .5,  .5}{60.2} & \textcolor[rgb]{ .5,  .5,  .5}{61.0} & \textcolor[rgb]{ .5,  .5,  .5}{2} \\
			& \textcolor[rgb]{ .5,  .5,  .5}{FairMOT~\cite{fairmot}} & \textcolor[rgb]{ .5,  .5,  .5}{40.2} & \textcolor[rgb]{ .5,  .5,  .5}{57.5} & \textcolor[rgb]{ .5,  .5,  .5}{53.3} & \textcolor[rgb]{ .5,  .5,  .5}{75.7} & \textcolor[rgb]{ .5,  .5,  .5}{28.2} & \textcolor[rgb]{ .5,  .5,  .5}{29.4} & \textcolor[rgb]{ .5,  .5,  .5}{78.3} & \textcolor[rgb]{ .5,  .5,  .5}{38.5} & \textcolor[rgb]{ .5,  .5,  .5}{56.0} & \textcolor[rgb]{ .5,  .5,  .5}{41.8} & \textcolor[rgb]{ .5,  .5,  .5}{3} \\
			\midrule
			\textcolor[rgb]{ .5,  .5,  .5}{YOLO-v5~\cite{yolov5}} & \textcolor[rgb]{ .5,  .5,  .5}{CSTrack~\cite{cstrack}} & \textcolor[rgb]{ .5,  .5,  .5}{34.2} & \textcolor[rgb]{ .5,  .5,  .5}{49.6} & \textcolor[rgb]{ .5,  .5,  .5}{47.0} & \textcolor[rgb]{ .5,  .5,  .5}{79.9} & \textcolor[rgb]{ .5,  .5,  .5}{23.7} & \textcolor[rgb]{ .5,  .5,  .5}{24.5} & \textcolor[rgb]{ .5,  .5,  .5}{81.1} & \textcolor[rgb]{ .5,  .5,  .5}{33.2} & \textcolor[rgb]{ .5,  .5,  .5}{46.7} & \textcolor[rgb]{ .5,  .5,  .5}{34.5} & \textcolor[rgb]{ .5,  .5,  .5}{4} \\
			\midrule
			\multirow{4}[2]{*}{\textcolor[rgb]{ .5,  .5,  .5}{Deformable DETR~\cite{deformabledetr}}} & \textcolor[rgb]{ .5,  .5,  .5}{TransTrack~\cite{transtrack}} & \textcolor[rgb]{ .5,  .5,  .5}{66.8} & \textcolor[rgb]{ .5,  .5,  .5}{73.9} & \textcolor[rgb]{ .5,  .5,  .5}{64.2} & \textcolor[rgb]{ .5,  .5,  .5}{76.9} & \textcolor[rgb]{ .5,  .5,  .5}{60.3} & \textcolor[rgb]{ .5,  .5,  .5}{64.6} & \textcolor[rgb]{ .5,  .5,  .5}{82.1} & \textcolor[rgb]{ .5,  .5,  .5}{62.1} & \textcolor[rgb]{ .5,  .5,  .5}{71.4} & \textcolor[rgb]{ .5,  .5,  .5}{71.4} & \textcolor[rgb]{ .5,  .5,  .5}{5} \\
			& \textcolor[rgb]{ .5,  .5,  .5}{TrackFormer~\cite{trackformer}} & \textcolor[rgb]{ .5,  .5,  .5}{67.4} & \textcolor[rgb]{ .5,  .5,  .5}{74.5} & \textcolor[rgb]{ .5,  .5,  .5}{66.0} & \textcolor[rgb]{ .5,  .5,  .5}{78.9} & \textcolor[rgb]{ .5,  .5,  .5}{61.1} & \textcolor[rgb]{ .5,  .5,  .5}{65.8} & \textcolor[rgb]{ .5,  .5,  .5}{82.8} & \textcolor[rgb]{ .5,  .5,  .5}{63.3} & \textcolor[rgb]{ .5,  .5,  .5}{74.1} & \textcolor[rgb]{ .5,  .5,  .5}{72.4} & \textcolor[rgb]{ .5,  .5,  .5}{6} \\
			& \textcolor[rgb]{ .5,  .5,  .5}{P3AFormer~\cite{p3aformer}} & \textcolor[rgb]{ .5,  .5,  .5}{42.3} & \textcolor[rgb]{ .5,  .5,  .5}{40.9} & \textcolor[rgb]{ .5,  .5,  .5}{38.1} & \textcolor[rgb]{ .5,  .5,  .5}{71.9} & \textcolor[rgb]{ .5,  .5,  .5}{44.0} & \textcolor[rgb]{ .5,  .5,  .5}{46.9} & \textcolor[rgb]{ .5,  .5,  .5}{75.7} & \textcolor[rgb]{ .5,  .5,  .5}{40.7} & \textcolor[rgb]{ .5,  .5,  .5}{43.8} & \textcolor[rgb]{ .5,  .5,  .5}{52.9} & \textcolor[rgb]{ .5,  .5,  .5}{7} \\
			& \textcolor[rgb]{ .5,  .5,  .5}{TransCenter~\cite{transcenter}} & \textcolor[rgb]{ .5,  .5,  .5}{63.5} & \textcolor[rgb]{ .5,  .5,  .5}{73.2} & \textcolor[rgb]{ .5,  .5,  .5}{65.8} & \textcolor[rgb]{ .5,  .5,  .5}{77.6} & \textcolor[rgb]{ .5,  .5,  .5}{55.3} & \textcolor[rgb]{ .5,  .5,  .5}{58.5} & \textcolor[rgb]{ .5,  .5,  .5}{82.8} & \textcolor[rgb]{ .5,  .5,  .5}{60.0} & \textcolor[rgb]{ .5,  .5,  .5}{76.4} & \textcolor[rgb]{ .5,  .5,  .5}{68.6} & \textcolor[rgb]{ .5,  .5,  .5}{8} \\
			\midrule
			\multirow{4}[2]{*}{\textcolor[rgb]{ .5,  .5,  .5}{YOLOX~\cite{ge2021yolox}}} & \textcolor[rgb]{ .5,  .5,  .5}{SORT~\cite{sort}} & \textcolor[rgb]{ .5,  .5,  .5}{63.2} & \textcolor[rgb]{ .5,  .5,  .5}{64.2} & \textcolor[rgb]{ .5,  .5,  .5}{60.6} & \textcolor[rgb]{ .5,  .5,  .5}{78.7} & \textcolor[rgb]{ .5,  .5,  .5}{62.3} & \textcolor[rgb]{ .5,  .5,  .5}{65.5} & \textcolor[rgb]{ .5,  .5,  .5}{82.0} & \textcolor[rgb]{ .5,  .5,  .5}{61.2} & \textcolor[rgb]{ .5,  .5,  .5}{75.5} & \textcolor[rgb]{ .5,  .5,  .5}{77.2} & \textcolor[rgb]{ .5,  .5,  .5}{9} \\
			& \textcolor[rgb]{ .5,  .5,  .5}{IOUTracker~\cite{ioutracker}} & \textcolor[rgb]{ .5,  .5,  .5}{\textbf{70.5}} & \textcolor[rgb]{ .5,  .5,  .5}{\textbf{75.2}} & \textcolor[rgb]{ .5,  .5,  .5}{\textbf{67.5}} & \textcolor[rgb]{ .5,  .5,  .5}{79.0} & \textcolor[rgb]{ .5,  .5,  .5}{66.3} & \textcolor[rgb]{ .5,  .5,  .5}{70.2} & \textcolor[rgb]{ .5,  .5,  .5}{85.5} & \textcolor[rgb]{ .5,  .5,  .5}{66.6} & \textcolor[rgb]{ .5,  .5,  .5}{\textbf{78.5}} & \textcolor[rgb]{ .5,  .5,  .5}{76.4} & \textcolor[rgb]{ .5,  .5,  .5}{10} \\
			& \textcolor[rgb]{ .5,  .5,  .5}{ByteTrack~\cite{bytetrack}} & \textcolor[rgb]{ .5,  .5,  .5}{65.2} & \textcolor[rgb]{ .5,  .5,  .5}{66.3} & \textcolor[rgb]{ .5,  .5,  .5}{61.2} & \textcolor[rgb]{ .5,  .5,  .5}{75.3} & \textcolor[rgb]{ .5,  .5,  .5}{64.1} & \textcolor[rgb]{ .5,  .5,  .5}{69.0} & \textcolor[rgb]{ .5,  .5,  .5}{77.1} & \textcolor[rgb]{ .5,  .5,  .5}{62.5} & \textcolor[rgb]{ .5,  .5,  .5}{77.2} & \textcolor[rgb]{ .5,  .5,  .5}{\textbf{82.3}} & \textcolor[rgb]{ .5,  .5,  .5}{11} \\
			& \textcolor[rgb]{ .5,  .5,  .5}{OC-SORT~\cite{ocsort}} & \textcolor[rgb]{ .5,  .5,  .5}{68.9} & \textcolor[rgb]{ .5,  .5,  .5}{69.2} & \textcolor[rgb]{ .5,  .5,  .5}{65.4} & \textcolor[rgb]{ .5,  .5,  .5}{\textbf{83.8}} & \textcolor[rgb]{ .5,  .5,  .5}{\textbf{68.7}} & \textcolor[rgb]{ .5,  .5,  .5}{\textbf{72.1}} & \textcolor[rgb]{ .5,  .5,  .5}{\textbf{86.8}} & \textcolor[rgb]{ .5,  .5,  .5}{\textbf{66.8}} & \textcolor[rgb]{ .5,  .5,  .5}{77.1} & \textcolor[rgb]{ .5,  .5,  .5}{79.3} & \textcolor[rgb]{ .5,  .5,  .5}{12} \\
			\midrule
			\multicolumn{13}{l}{\textit{b) Zero-shot setting}} \\
			\midrule
			\multirow{10}[2]{*}{YOLOX~\cite{ge2021yolox}} & SORT~\cite{sort}  & 54.2  & 55.4  & 52.2  & 79.5  & 53.0  & 55.7  & 82.8  & 52.5  & 60.6  & 63.6  & 14 \\
			& IOUTracker~\cite{ioutracker} & 55.6  & 57.3  & 54.1  & 84.9  & 53.9  & 57.4  & 84.1  & 53.9  & 60.1  & 59.1  & 15 \\
			& DeepSORT~\cite{deepsort} & 44.7  & 53.1  & 48.1  & 71.0  & 37.8  & 40.4  & 74.8  & 42.3  & 51.3  & 49.9  & 16 \\
			& Tracktor++~\cite{tracktor++} & 29.0  & 65.8  & 60.9  & 82.4  & 12.9  & 20.4  & 29.7  & 27.8  & 35.0  & 26.2  & 17 \\
			& ByteTrack~\cite{bytetrack} & 54.8  & 56.0  & 51.6  & 73.3  & 53.7  & 58.5  & 73.4  & 52.5  & 61.5  & \textbf{68.4} & 18 \\
			& OC-SORT~\cite{ocsort} & 58.5  & 57.7  & 55.2  & \textbf{87.7} & \textbf{59.4} & \textbf{62.0} & \textbf{87.9} & 57.2  & 61.0  & 66.6  & 19 \\
			& StrongSORT~\cite{strongsort} & 43.2  & 54.7  & 48.3  & 73.0  & 34.2  & 36.8  & 74.2  & 40.4  & 47.9  & 43.4  & 20 \\
			& StrongSORT++~\cite{strongsort} & 42.9  & 54.1  & 44.2  & 61.7  & 34.4  & 37.5  & 69.2  & 38.6  & 39.4  & 42.9  & 21 \\
			& Deep OC-SORT~\cite{deepocsort} & 33.6  & 26.4  & 25.4  & 81.5  & 42.8  & 45.3  & 84.3  & 32.9  & 25.7  & 39.3  & 22 \\
			& \cellcolor[rgb]{ .906,  .902,  .902}\textbf{NetTrack} (ours) & \cellcolor[rgb]{ .906,  .902,  .902}\textbf{63.3} & \cellcolor[rgb]{ .906,  .902,  .902}\textbf{70.6} & \cellcolor[rgb]{ .906,  .902,  .902}\textbf{61.2} & \cellcolor[rgb]{ .906,  .902,  .902}77.6 & \cellcolor[rgb]{ .906,  .902,  .902}56.8 & \cellcolor[rgb]{ .906,  .902,  .902}60.5 & \cellcolor[rgb]{ .906,  .902,  .902}83.3 & \cellcolor[rgb]{ .906,  .902,  .902}\textbf{58.8} & \cellcolor[rgb]{ .906,  .902,  .902}\textbf{62.5} & \cellcolor[rgb]{ .906,  .902,  .902}65.1 & \cellcolor[rgb]{ .906,  .902,  .902}23 \\
			\midrule
			\multirow{5}[2]{*}{Grounding DINO~\cite{groundingdino}} & SORT~\cite{sort}  & 59.9  & 63.9  & 60.1  & 81.1  & 56.2  & 58.9  & 84.1  & 57.9  & 71.4  & 69.7  & 24 \\
			& IOUTracker~\cite{ioutracker} & 70.9  & 77.4  & 62.3  & 71.9  & 65.0  & \textbf{70.8} & 82.4  & 63.5  & 65.8  & 70.7  & 25 \\
			& ByteTrack~\cite{bytetrack} & 64.1  & 67.9  & 61.1  & 73.0  & 60.5  & 66.5  & 73.7  & 60.7  & 74.9  & \textbf{78.9} & 26 \\
			& OC-SORT~\cite{ocsort} & 69.0  & 70.9  & 66.8  & \textbf{87.9} & \textbf{67.2} & 70.1  & \textbf{90.4} & 66.9  & 73.6  & 76.0  & 27 \\
			& \cellcolor[rgb]{ .906,  .902,  .902}\textbf{NetTrack} (ours) & \cellcolor[rgb]{ .906,  .902,  .902}\textbf{72.5} & \cellcolor[rgb]{ .906,  .902,  .902}\textbf{80.7} & \cellcolor[rgb]{ .906,  .902,  .902}\textbf{72.6} & \cellcolor[rgb]{ .906,  .902,  .902}83.3 & \cellcolor[rgb]{ .906,  .902,  .902}65.2 & \cellcolor[rgb]{ .906,  .902,  .902}70.4 & \cellcolor[rgb]{ .906,  .902,  .902}82.5 & \cellcolor[rgb]{ .906,  .902,  .902}\textbf{68.7} & \cellcolor[rgb]{ .906,  .902,  .902}\textbf{78.9} & \cellcolor[rgb]{ .906,  .902,  .902}77.0 & \cellcolor[rgb]{ .906,  .902,  .902}28 \\
			\bottomrule
		\end{tabular}%
	}
	\label{tab:bft}%
	\vspace{-10pt}
\end{table*}%

The experimental section aims to validate the following core conclusions of this work:
\begin{enumerate}
    \item[1)] High dynamicity of open-world objects poses significant challenges for MOT.
    \item[2)] NetTrack outstandingly handles dynamic objects and exhibits strong generalization abilities on diverse open-world tracking datasets without finetuning.
    \item[3)] The proposed fine-grained learning shows stronger generalization abilities for tracking dynamic objects compared to coarse-grained methods.
\end{enumerate}

\subsection{Settings}
\noindent\textbf{Dataset}~
BFT is utilized to assess the performance of trackers in highly dynamic open-world scenarios. In zero-shot transfer evaluation, the validation sets of the large-scale TAO-OW~\cite{tao-ow} and TAO~\cite{dave2020tao} are employed for extensive generalization ability assessment. Specifically, the evaluation of TAO follows the description in \cite{ovtrack}, where an open-vocabulary setting is adopted for \textit{base} and \textit{novel} categories, and the classification ability of trackers is evaluated. \textit{Novel} classes are the classes defined as rare in the LVIS~\cite{lvis} dataset. Differently, object classes of TAO-OW are divided into \textit{known} and \textit{unknown} based on whether they belong to the 80 categories in COCO~\cite{coco}. 
In the ablation experiments, in addition to TAO and TAO-OW, AnimalTrack~\cite{zhang2022animaltrack} and GMOT-40~\cite{bai2021gmot} are also included as references and evaluated in an open-world setting following TAO-OW. Regarding AnimalTrack, 8 out of 10 classes are outside the COCO categories. Similarly, 12 out of 18 classes in GMOT-40 are outside the COCO categories. 

\noindent\textbf{Metrics}~
Open-world tracking accuracy (OWTA)~\cite{tao-ow} is an open-world MOT metric proposed for TAO-OW and is the main metric in our experiments. OWTA evaluates both \textit{detection recall} (D. Re.) and \textit{association accuracy} (A. Acc.), respectively. 
\textit{Detection accuracy} (D. Acc.), \textit{detection precision} (D. Pr.), \textit{association recall} (A. Re.), and \textit{ association precision} (A. Pr.) are reference metrics.
TETA~\cite{tetr} aims to evaluate multi-category objects and is used to evaluate the TAO dataset under an open-vocabulary setting. \textit{Localization score} (LocA) and \textit{association score} (AssocA) are calculated in TETA. 
HOTA~\cite{HOTA}, MOTA~\cite{clear}, and IDF1~\cite{idf1} are classic metrics used for comparisons with classic MOT methods on BFT and serve as references. All evaluation processes are adopted from TrackEval~\cite{trackeval}.

\noindent\textbf{Implementation details}~
In NetTrack, the coarse-grained association adapts from BYTE~\cite{bytetrack}, and the default point tracker adapts from CoTracker~\cite{cotracker} pretrained on TAP-Vid-Kubric~\cite{tapvid}. By default, the tracking stride is 8, lost tracks are retained for 30 frames, and the initialized point sampling is with a grid of \texttt{(3,3)}. The default detector is GroundingDINO~\cite{groundingdino} with Swin-B~\cite{swin} backbone, which was pre-trained on COCO~\cite{coco}, O365~\cite{o365}, \etc. To validate the generalization ability of NetTrack in an affordable manner for open-world MOT applications, no additional training is required for all evaluated benchmarks. The finetuning and evaluation of the publicly available SoTA trackers on BFT followed their default settings.

\subsection{High Dynamicity Evaluation}

\begin{table}[t]
  \centering
  \caption{Zero-shot transfer evaluation on open-vocabulary MOT comparison. $^\star$ denotes comprehensive evaluation metrics, and $^*$ represents non open-world setting, \textit{i.e.}, also trained on \textit{novel} classes on TAO. Results of finetuning and learning \textit{novel} classes are shown in \textcolor[rgb]{0.5, 0.5, 0.5}{gray}, and the best results are shown in \textbf{bold}.}
    \resizebox{\linewidth}{!}{
    \begin{tabular}{ccccccccc}
    \toprule
    \multirow{3}[6]{*}{\textbf{Method}} & \multicolumn{8}{c}{\textbf{TAO benchmark  evaluation}} \\
\cmidrule{2-9}          & \multicolumn{4}{c}{\textbf{Base}} & \multicolumn{4}{c}{\textbf{Novel}} \\
\cmidrule(r){2-5}  \cmidrule(r){6-9}         & TETA$^\star$↑ & LocA↑ & AssocA↑ & ClsA↑ & TETA$^\star$↑ & LocA↑ & AssocA↑ & ClsA↑ \\
    \midrule
    \multicolumn{9}{l}{\textcolor[rgb]{0.5, 0.5, 0.5}{\textit{a) Finetuned on TAO dataset}}} \\
    \midrule
    \textcolor[rgb]{0.5, 0.5, 0.5}{DeepSORT~\cite{deepsort}} & \textcolor[rgb]{0.5, 0.5, 0.5}{26.9} & \textcolor[rgb]{0.5, 0.5, 0.5}{47.1} & \textcolor[rgb]{0.5, 0.5, 0.5}{15.8} & \textcolor[rgb]{0.5, 0.5, 0.5}{17.7} & \textcolor[rgb]{0.5, 0.5, 0.5}{21.1} & \textcolor[rgb]{0.5, 0.5, 0.5}{46.4} & \textcolor[rgb]{0.5, 0.5, 0.5}{14.7} & \textcolor[rgb]{0.5, 0.5, 0.5}{2.3} \\
    \textcolor[rgb]{0.5, 0.5, 0.5}{Tracktor++~\cite{tracktor++}} & \textcolor[rgb]{0.5, 0.5, 0.5}{28.3} & \textcolor[rgb]{0.5, 0.5, 0.5}{47.4} & \textcolor[rgb]{0.5, 0.5, 0.5}{20.5} & \textcolor[rgb]{0.5, 0.5, 0.5}{17.0} & \textcolor[rgb]{0.5, 0.5, 0.5}{22.7} & \textcolor[rgb]{0.5, 0.5, 0.5}{46.7} & \textcolor[rgb]{0.5, 0.5, 0.5}{19.3} & \textcolor[rgb]{0.5, 0.5, 0.5}{2.2} \\
    \textcolor[rgb]{0.5, 0.5, 0.5}{QDTrack$^*$~\cite{qdtrack}} & \textcolor[rgb]{0.5, 0.5, 0.5}{27.1} & \textcolor[rgb]{0.5, 0.5, 0.5}{45.6} & \textcolor[rgb]{0.5, 0.5, 0.5}{24.7} & \textcolor[rgb]{0.5, 0.5, 0.5}{11.0} & \textcolor[rgb]{0.5, 0.5, 0.5}{22.5} & \textcolor[rgb]{0.5, 0.5, 0.5}{42.7} & \textcolor[rgb]{0.5, 0.5, 0.5}{24.4} & \textcolor[rgb]{0.5, 0.5, 0.5}{0.4} \\
    \textcolor[rgb]{0.5, 0.5, 0.5}{TETer$^*$~\cite{tetr}} & \textcolor[rgb]{0.5, 0.5, 0.5}{30.3} & \textcolor[rgb]{0.5, 0.5, 0.5}{47.4} & \textcolor[rgb]{0.5, 0.5, 0.5}{31.6} & \textcolor[rgb]{0.5, 0.5, 0.5}{12.1} & \textcolor[rgb]{0.5, 0.5, 0.5}{25.7} & \textcolor[rgb]{0.5, 0.5, 0.5}{45.9} & \textcolor[rgb]{0.5, 0.5, 0.5}{31.1} & \textcolor[rgb]{0.5, 0.5, 0.5}{0.2} \\
    \midrule
    \multicolumn{9}{l}{\textcolor[rgb]{0.5, 0.5, 0.5}{\textit{b) Trained with LVIS dataset}}} \\
    \midrule
    \textcolor[rgb]{0.5, 0.5, 0.5}{OVTrack}~\cite{ovtrack}& \textcolor[rgb]{0.5, 0.5, 0.5}{\textbf{{35.5}}} & \textcolor[rgb]{0.5, 0.5, 0.5}{\textbf{{49.3}}} & \textcolor[rgb]{0.5, 0.5, 0.5}{\textbf{{36.9}}} & \textcolor[rgb]{0.5, 0.5, 0.5}{20.2} & \textcolor[rgb]{0.5, 0.5, 0.5}{27.8} & \textcolor[rgb]{0.5, 0.5, 0.5}{48.8} & \textcolor[rgb]{0.5, 0.5, 0.5}{\textbf{{33.6}}} & \textcolor[rgb]{0.5, 0.5, 0.5}{1.5} \\
    \midrule
    \multicolumn{9}{l}{\textit{c) Zero-shot setting}} \\
    \midrule
    \rowcolor[rgb]{ .906,  .902,  .902} \textbf{NetTrack} (ours) &    33.0  &   45.7    &   28.6    &    \textbf{24.8}   &    \textbf{32.6}   &  \textbf{51.3}      &    33.0   &  \textbf{13.3} \\
    \bottomrule
    \end{tabular}}%
  \label{tab:tao_teta}
\end{table}

\begin{table}[t]
  \centering
  \caption{Zero-shot transfer evaluation on TAO-OW. $^\star$ denotes comprehensive evaluation metrics, and $^*$ represents non open-world setting, \textit{i.e.}, also trained on \textit{unknown} classes. Finetuned results are in \textcolor[rgb]{0.5, 0.5, 0.5}{gray}, and the best results are shown in \textbf{bold}.}
  \resizebox{\linewidth}{!}{
    \begin{tabular}{lcccccc}
    \toprule
    \multicolumn{1}{c}{\multirow{3}[6]{*}{\textbf{Method}}} & \multicolumn{6}{c}{\textbf{TAO-OW benchmark evaluation}} \\
\cmidrule{2-7}          & \multicolumn{3}{c}{\textbf{Known}} & \multicolumn{3}{c}{\textbf{Unknown}} \\
\cmidrule(r){2-4}  \cmidrule(r){5-7}         & OWTA$^\star$↑ & D. Re.↑ & A. Acc.↑ & OWTA$^\star$↑ & D. Re.↑ & A. Acc.↑ \\
    \midrule
    \multicolumn{7}{l}{\textcolor[rgb]{0.5, 0.5, 0.5}{\textit{a) Finetuned on TAO-OW dataset}}} \\
    \midrule
    \textcolor[rgb]{0.5, 0.5, 0.5}{SORT}~\cite{sort} & \textcolor[rgb]{0.5, 0.5, 0.5}{46.6} & \textcolor[rgb]{0.5, 0.5, 0.5}{67.4} & \textcolor[rgb]{0.5, 0.5, 0.5}{33.7} & \textcolor[rgb]{0.5, 0.5, 0.5}{33.9} & \textcolor[rgb]{0.5, 0.5, 0.5}{43.4} & \textcolor[rgb]{0.5, 0.5, 0.5}{30.3} \\
    \textcolor[rgb]{0.5, 0.5, 0.5}{SORT-TAO$^*$}~\cite{dave2020tao} & \textcolor[rgb]{0.5, 0.5, 0.5}{54.2} & \textcolor[rgb]{0.5, 0.5, 0.5}{74.0} & \textcolor[rgb]{0.5, 0.5, 0.5}{40.6} & \textcolor[rgb]{0.5, 0.5, 0.5}{39.9} & \textcolor[rgb]{0.5, 0.5, 0.5}{68.8} & \textcolor[rgb]{0.5, 0.5, 0.5}{24.1} \\
    \textcolor[rgb]{0.5, 0.5, 0.5}{AOA$^*$}~\cite{aoa} & \textcolor[rgb]{0.5, 0.5, 0.5}{52.8} & \textcolor[rgb]{0.5, 0.5, 0.5}{72.5} & \textcolor[rgb]{0.5, 0.5, 0.5}{39.1} & \textcolor[rgb]{0.5, 0.5, 0.5}{49.7} & \textcolor[rgb]{0.5, 0.5, 0.5}{74.7} & \textcolor[rgb]{0.5, 0.5, 0.5}{33.4} \\
    \textcolor[rgb]{0.5, 0.5, 0.5}{Tracktor}~\cite{tracktor++} & \textcolor[rgb]{0.5, 0.5, 0.5}{57.9} & \textcolor[rgb]{0.5, 0.5, 0.5}{\textbf{80.2}} & \textcolor[rgb]{0.5, 0.5, 0.5}{42.6} & \textcolor[rgb]{0.5, 0.5, 0.5}{22.8} & \textcolor[rgb]{0.5, 0.5, 0.5}{54.0} & \textcolor[rgb]{0.5, 0.5, 0.5}{10.0} \\
    \textcolor[rgb]{0.5, 0.5, 0.5}{OWTB}~\cite{tao-ow} & \textcolor[rgb]{0.5, 0.5, 0.5}{60.2} & \textcolor[rgb]{0.5, 0.5, 0.5}{77.2} & \textcolor[rgb]{0.5, 0.5, 0.5}{47.4} & \textcolor[rgb]{0.5, 0.5, 0.5}{39.2} & \textcolor[rgb]{0.5, 0.5, 0.5}{46.9} & \textcolor[rgb]{0.5, 0.5, 0.5}{34.5} \\
    \textcolor[rgb]{0.5, 0.5, 0.5}{Video OWL-ViT}~\cite{video-owlvit} & \textcolor[rgb]{0.5, 0.5, 0.5}{59.0} & \textcolor[rgb]{0.5, 0.5, 0.5}{69.0} & \textcolor[rgb]{0.5, 0.5, 0.5}{\textbf{51.5}} & \textcolor[rgb]{0.5, 0.5, 0.5}{\textbf{45.4}} & \textcolor[rgb]{0.5, 0.5, 0.5}{53.4} & \textcolor[rgb]{0.5, 0.5, 0.5}{\textbf{40.5}} \\
    \midrule
    \multicolumn{7}{l}{\textit{b) Zero-shot setting}} \\
    \midrule
    \rowcolor[rgb]{ .906,  .902,  .902} \textbf{NetTrack} (ours) & \textbf{62.7} & 77.4  & 51.0 & 43.7 & \textbf{58.7} & 33.2 \\
    \bottomrule
    \end{tabular}}%
  \label{tab:tao-ow}%
  \vspace{-5pt}
\end{table}%

\begin{table*}[t]
  \centering
  \caption{Association comparison between fine-grained Nets and solely coarse-grained methods on TAO and TAO-OW validation benchmark in an open-vocabulary~\cite{ovtrack} and an open-world setting~\cite{tao-ow}, respectively. The best results are shown in \textbf{bold}. The same  detector~\cite{groundingdino} is used.}
  \resizebox{0.9\linewidth}{!}{
    \begin{tabular}{ccccccccccccccc}
    \toprule
    \multirow{3}[6]{*}{\textbf{Method}} & \multicolumn{8}{c}{\textbf{TAO benchmark  evaluation}}        & \multicolumn{6}{c}{\textbf{TAO-OW benchmark evaluation}} \\
\cmidrule(r){2-9} \cmidrule(r){10-15}          & \multicolumn{4}{c}{\textbf{Base}} & \multicolumn{4}{c}{\textbf{Novel}} & \multicolumn{3}{c}{\textbf{Known}} & \multicolumn{3}{c}{\textbf{Unknown}} \\
\cmidrule(r){2-5} \cmidrule(r){6-9} \cmidrule(r){10-12} \cmidrule(r){13-15}         & TETA$^\star$↑ & LocA↑ & AssocA↑ & ClsA↑ & TETA$^\star$↑ & LocA↑ & AssocA↑ & ClsA↑ & OWTA$^\star$↑ & D. Re.↑ & A. Acc.↑ & OWTA$^\star$↑ & D. Re.↑ & A. Acc.↑ \\
    \midrule
    SORT~\cite{sort}  & 28.5  & 33.8  & 27.3  & 24.5  & 28.1  & 41.7  & 28.6  & \textbf{13.7} & 62.4  & 72.8  & 53.8  & 39.5  & 39.6  & 40.6 \\
    ByteTrack~\cite{bytetrack} & 32.6  & 41.7  & \textbf{31.2} & 24.7  & 32.1  & 48.8  & \textbf{34.4} & 13.3  & \textbf{63.3} & 71.3  & 56.4  & 40.9  & 40.5  & \textbf{42.5} \\
    OC-SORT~\cite{ocsort} & 25.8  & 32.4  & 20.5  & 24.5  & 25.7  & 39.0  & 24.7  & 13.4  & 48.7  & 69.0  & 34.4  & 31.6  & 37.8  & 27.1 \\
    \midrule
    \textbf{NetTrack} (ours) & \textbf{33.0} & \textbf{45.7} & 28.6  & \textbf{24.8} & \textbf{32.6} & \textbf{51.3} & 33.0  & 13.3  & 62.7  & \textbf{77.4} & 51.0  & \textbf{43.7} & \textbf{58.7} & 33.2 \\
    \bottomrule
    \end{tabular}%
    }%
  \label{tab:abla1}%
\vspace{-5pt}
\end{table*}%

\begin{table*}[t]
  \centering
    \caption{Association comparison between fine-grained Nets and solely coarse-grained methods on AnimalTrack and GMOT-40 benchmark following the open-world setting of TAO-OW~\cite{tao-ow} for reference. The best results are shown in \textbf{bold}. The same detector~\cite{groundingdino} is used.}
  \resizebox{0.8\linewidth}{!}{
    \begin{tabular}{ccccccccccccc}
    \toprule
    \multirow{3}[6]{*}{\textbf{Method}} & \multicolumn{6}{c}{\textbf{AnimalTrack benchmark evaluation}} & \multicolumn{6}{c}{\textbf{GMOT-40 benchmark evaluation}} \\
\cmidrule(r){2-7} \cmidrule(r){8-13}           & \multicolumn{3}{c}{\textbf{Known}} & \multicolumn{3}{c}{\textbf{Unknown}} & \multicolumn{3}{c}{\textbf{Known}} & \multicolumn{3}{c}{\textbf{Unknown}} \\
\cmidrule(r){2-4} \cmidrule(r){5-7} \cmidrule(r){8-10} \cmidrule(r){11-13}          & OWTA$^\star$↑ & D. Re.↑ & A. Acc.↑ & OWTA$^\star$↑ & D. Re.↑ & A. Acc.↑ & OWTA$^\star$↑ & D. Re.↑ & A. Acc.↑ & OWTA$^\star$↑ & D. Re.↑ & A. Acc.↑ \\
    \midrule
    SORT~\cite{sort}  & 44.2  & 32.7  & 60.0  & 48.1  & 42.6  & 54.7  & 43.6  & 30.6  & 62.4  & 35.8  & 24.1  & 53.4 \\
    ByteTrack~\cite{bytetrack} & 41.7  & 28.1  & 62.2  & 46.7  & 41.7  & 52.6  & 41.3  & 27.2  & \textbf{63.1} & 35.8  & 21.8  & \textbf{59.2} \\
    OC-SORT~\cite{ocsort} & 45.2  & 32.9  & \textbf{62.6} & 48.7  & 43.0  & \textbf{55.8} & 44.0  & 31.0  & 62.7  & 36.4  & 24.6  & 54.1 \\
    \midrule
    \textbf{NetTrack} (ours) & \textbf{48.1} & \textbf{42.0} & 55.6  & \textbf{51.4} & \textbf{50.8} & 52.5  & \textbf{45.9} & \textbf{37.8} & 56.1  & \textbf{36.6} & \textbf{29.3} & 45.7 \\
    \bottomrule
    \end{tabular}%
    }%
\label{tab:abla2}%
\vspace{-7pt}
\end{table*}%

A comprehensive evaluation of NetTrack and other SoTA trackers on the highly dynamic BFT is presented in \cref{tab:bft}. 
The evaluation is divided into two main parts: \textit{a)} Fine-tuning on the BFT dataset using closed-set trackers. 
\textit{b)} Open-world MOT condition, which involves tracking under zero-shot settings. 
To ensure a fair evaluation of tracker performance in the highly dynamic challenges of the open-world scenarios,  all text prompts for open-world conditions only include \texttt{'bird'}, consistent with the category in the COCO dataset that is used to train closed-set trackers.
The experimental results mainly demonstrate that: 
\begin{enumerate}
\item[1)] Even in the zero-shot open-world tracking setting, NetTrack achieves superior performance compared to SoTA finetuned closed-set trackers. NetTrack improves 1.3 points on OWTA compared with the best finetuned results, confirming the zero-shot generalization ability of the proposed framework.
\item[2)] In comparison to the results after fine-tuning (lines 9-12), closed-set trackers exhibit sub-optimal zero-shot generalization ability (lines 13,14,17,18) in highly dynamic open-world scenarios, with an average decrease of 16\% on OWTA, 15\% on HOTA, and 21\% on MOTA, which indicates closed-set trackers have suboptimal generalization ability on dynamic objects in the open world.
\item[3)] NetTrack encourages associating potential objects of interest and achieves an improvement on \textit{detection recall} by 3.4 points. It also results in more false positive samples and adds pressure to the association with a slight decrease in A. Acc. However the comprehensive OWTA gets promoted by 1.6 points compared with the best coarse-grained association methods (lines 24-27).

\end{enumerate}

\subsection{Zero-Shot Transfer Evaluation}

\noindent\textbf{Zero-shot transfer on open-vocabulary settings}~
In \cref{tab:tao_teta}, zero-shot transfer on TAO with open-vocabulary MOT evaluation is shown. DeepSORT~\cite{deepsort} and Tracktor++~\cite{tracktor++} are with ViLD~\cite{vild} as the detector. OVTrack~\cite{ovtrack} is trained on a generated dataset derived from LVIS~\cite{lvis}, which exhibits a high level of class consistency with TAO. Compared to finetuned trackers, NetTrack significantly improves tracking classification accuracy and achieves strong zero-shot tracking accuracy. Although NetTrack is susceptible to a large number of false positive samples due to the absence of finetuning, which puts it at a slight disadvantage in the evaluation of LocA and AssocA in the \textit{base} classes, the proposed framework achieves an 11.8-point increase in ClsA, a 2.5-point increase in LocA, comparable AssocA in the \textit{novel} classes and a 4.5-point increase in overall TETA, further demonstrating its competitive generalization ability.

\noindent\textbf{Zero-shot transfer on open-world settings}~
Zero-shot generalization of NetTrack on the TAO-OW~\cite{tao-ow} benchmark is demonstrated in \cref{tab:tao-ow}. 
Apart from NetTrack, all trackers underwent fine-tuning on \textit{known} classes of TAO-OW training set. Compared to finetuned SoTA trackers, NetTrack achieves optimal performance on \textit{known} classes. With the D. Re. being similar to the open world tracking baseline (OWTB)~\cite{tao-ow}, A. Acc. surpasses the baseline by 3.6 points, confirming the generalization ability of dynamicity-aware association. Similarly, while A. Acc. remains approximate to Video OWL-ViT~\cite{video-owlvit}, D. Re. shows an improvement of 8.4 points, validating the effectiveness of fine-grained localization. On \textit{unknown} classes, the introduction of false positive samples leads to a slight decrease in A. Acc., but the overall OWTA performance is still competitive with a 5.3-point improvement on D. Re. 
\begin{figure}[t]
	\begin{center}
		\includegraphics[width=0.475\textwidth]{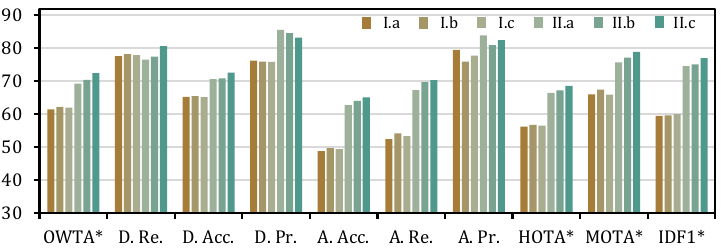}
	\end{center}
         \vspace{-10pt}
	\caption{Comparison of detachable modules in the proposed framework, where the SoTA grounding-based detectors~\cite{glip,groundingdino} as I, II and point trackers~\cite{pips,tapir,cotracker} as a, b, c are considered. The robust performance in module variations confirms the excellent generality of the proposed framework.}
	\label{fig:detach}
        \vspace{-10pt}
\end{figure}

\subsection{Ablations}
\label{sec:ablation}
\noindent\textbf{Generality of fine-grained Nets}~In \cref{tab:abla1} and \cref{tab:abla2}, the comparison between the proposed association with fine-grained Nets and coarse-grained methods~\cite{sort,bytetrack,ocsort} on TAO~\cite{dave2020tao}, TAO-OW~\cite{tao-ow}, AnimalTrack~\cite{zhang2022animaltrack}, and GMOT-40~\cite{bai2021gmot} are shown. Attributed to the proposed framework that encourages the discovery of more potential objects in open-world scenarios, NetTrack achieves significant improvements in LocA and D. Re. on both \textit{seen} and \textit{unseen} classes across four benchmarks. Particularly, the D. Re. of \textit{unknown} classes on TAO-OW exhibits a remarkable increase of 18.2 points compared to the second-best performance, confirming its strong generalization. Although the introduction of false positive samples leads to a slight decrease in AssoA and A. Acc, the overall TETA and OWTA have been significantly improved in both the \textit{seen} and \textit{unseen} classes.

\begin{figure}[t]
	\begin{center}
		\includegraphics[width=0.475\textwidth]{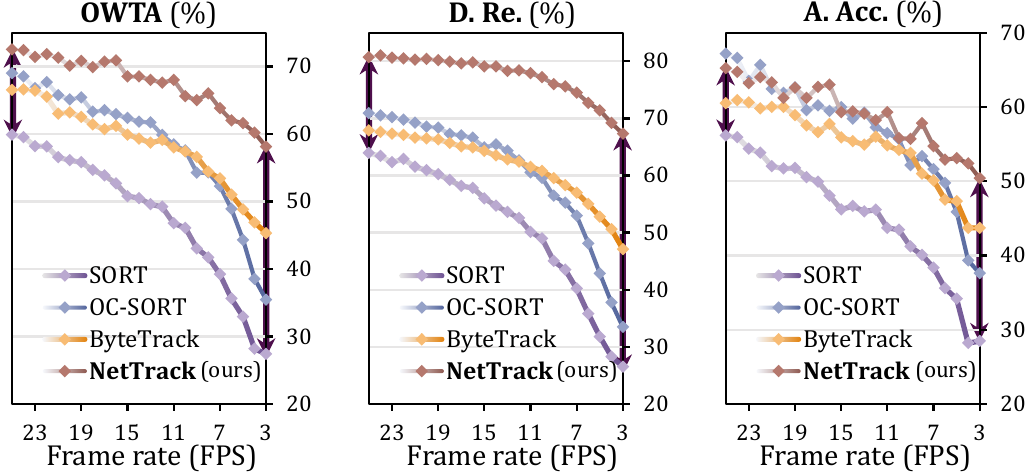}
	\end{center}
         \vspace{-10pt}
	\caption{Stability comparison against frame-rate drops between NetTrack and coarse-grained methods~\cite{sort, ocsort, bytetrack} on BFT. Better stability indicates stronger generalization ability of NetTrack.}
	\label{fig:frame_rate}
 \vspace{-15pt}
\end{figure}

\noindent\textbf{Robust framework with detachable modules}~To validate the generality of the proposed framework, \cref{fig:detach} shows ablation study on detachable modules, including open-set localization methods and point trackers. Specifically, the localization methods are denoted as GLIP~\cite{glip} I, Grounding DINO~\cite{groundingdino} II, and point trackers are denoted as PIPs~\cite{pips} a, TAPIR~\cite{tapir} b, CoTracker~\cite{cotracker} c. The combination of Grounding DINO and CoTracker is denoted as II.c and serves as the default setting.
In comparing localization ability, both methods demonstrate competitiveness in terms of D. Re. but~\cite{glip} exhibits a slight performance deficit in A. Acc and overall OWTA due to the introduction of more false positives. Similarly, three point trackers exhibit approximately excellent performance. Overall, the change of modules does not significantly degrade the overall performance, thus verifying the good generalization ability of the proposed framework.

\noindent\textbf{Stability against frame-rate drops}~In practical applications of open-world tracking, especially in scenarios related to edge devices~\cite{edge}, it is common to encounter reduced video frame rates due to the need to reduce computational load or save energy, further exacerbating the challenges posed by the dynamicity of open-world objects. \cref{fig:frame_rate} shows the tracking performance on the BFT dataset under reduced frame rates, from default frame rates (25 FPS) to one-tenth (3 FPS). Compared to other association methods~\cite{sort,ocsort,bytetrack}, NetTrack demonstrates good stability in the face of reduced frame rates. This further illustrates the generalization performance of the proposed framework. 

\noindent\textbf{\textit{Remark}}
~~
The qualitative tracking results of diverse scenarios are shown in supplemented \cref{sec:qualitative_supp}. The promising applications of NetTrack in video editing, open-world ecological inspection, and embedding descriptors for professional use are discussed in \cref{sec:app}.

\section{Conclusion}
This work focuses on the high dynamicity in open-world MOT and proposes NetTrack to learn fine-grained object cues. Specifically, fine-grained visual cues and object-text correspondence are introduced for dynamicity-aware association and localization.
This work also proposes a highly dynamic open-world MOT benchmark, BFT, and extensive evaluation with SoTA trackers proves the effectiveness of the proposed NetTrack for tracking dynamic objects. Moreover, extensive transfer experiments on several challenging open-world MOT benchmarks validate the strong generalization ability of NetTrack without finetuning. The analysis of limitations suggests that a more streamlined end-to-end manner and filtering false positive samples are promising for further improvement.
\section*{Acknowledge}  
This project is supported by the Innovation and Technology Commission of the HKSAR Government under the InnoHK initiative, ITF GHP/126/21GD, and HKU's CRF seed grant. We also appreciate precious biological advice from Ming-Shan Wang.
{
    \small
    \bibliographystyle{ieeenat_fullname}
    \bibliography{main}
}

\clearpage
\setcounter{page}{1}
\maketitlesupplementary
In the supplementary materials, we provide a detailed exposition of the NetTrack pipeline in \cref{sec:method_details}, statistics and elaborate dynamicity descriptions of the proposed BFT benchmark in \cref{sec:benchmak_details}, as well as a more comprehensive introduction to experimental details and more enriched experimental results in \cref{sec:exp_details}. Additionally, we also demonstrate three potential application scenarios of NetTrack in \cref{sec:app}.

\section{Method Details}\label{sec:method_details}
In contrast to previous methods~\cite{sort,bytetrack,ocsort} that utilize coarse-grained representations, NetTrack introduces a more fine-grained Net by incorporating fine-grained object information for tracking. The brief workflow of the proposed fine-grained Net is illustrated in \cref{algorithm}. Specifically, after initialization, NetTrack employs a simple and fast coarse tracker to predict the coarse-grained object trajectory $\mathcal{T}_\mathrm{o}^\mathrm{coarse}$. Subsequently, when the frame count reaches a tracking stride $\mathcal{S}$, NetTrack performs fine-grained sampling within this stride, obtaining the points of interest (POIs) $\mathcal{Q}$ and their corresponding trajectories. In fine-grained tracking, the fine-grained object trajectory is obtained by matching with the trajectories of POIs and the coarse-grained object trajectory, which is then used as the output.


\begin{algorithm}[t]
\caption{Pseudo code of fine-grained Net.}
\label{algorithm}
\DontPrintSemicolon
\footnotesize{
\KwIn{Video sequence $\mathbf{V}$, detector \normalsize{\texttt{D}}, \footnotesize 
physical point tracker {\normalsize{\texttt{Tr}}\footnotesize$_\mathrm{p}$}, tracking stride $\mathcal{S}$}
\KwOut{Object tracks $\mathcal{T}_\mathrm{o}$}
\BlankLine

{$\mathcal{T}_\mathrm{o}^\mathrm{coarse}, \mathcal{T}_\mathrm{o}^\mathrm{fine} \leftarrow \emptyset, \emptyset$ 
\scriptsize{{\textcolor[rgb]{0.5, 0.5, 0.5}{\texttt{\# Initialized object tracks}}}}} \;
	\For{$\mathbf{I}_k$ in $\mathbf{V}$}{

	{$\mathcal{D}_k \leftarrow$ {\normalsize{\texttt{D}}}($\mathbf{I}_k$)}\;
	{$\mathcal{T}_\mathrm{o}^\mathrm{coarse} \leftarrow {\normalsize{\texttt{Tr}}}_\mathrm{o}^\mathrm{coarse}(\mathcal{D}_k, \mathcal{T}_\mathrm{o}^\mathrm{coarse})$}\;
	{$\mathbf{I}_\mathrm{s}, \mathbf{D}_\mathrm{s} \leftarrow \mathbf{I}_\mathrm{s} \cup \{\mathrm{I}_k\}, \mathbf{D}_\mathrm{s} \cup \{\mathcal{D}_k\}$}\;
	{$N \leftarrow$ number of images in $\mathbf{I}_\mathrm{s}$}\;
         \If{ N == $\mathcal{S}$}
	{
	{\textcolor[rgb]{0.5, 0.5, 0.5}{\texttt{/* Step 1: Fine-grained sampling */}}}\;
		{$\mathcal{Q} \leftarrow$ \footnotesize{\texttt{Sampling}}($\mathcal{T}_o^\mathrm{coarse}$)} \;
		{$\mathcal{T}_\mathrm{p} \leftarrow$ {\texttt{Tr}$_\mathrm{p}$}($\mathbf{I}_\mathrm{s}, \mathcal{Q}$)} \;
        {\textcolor[rgb]{0.5, 0.5, 0.5}{\texttt{/* Step 2: Fine-grained matching */}}}\;
		\For{$\mathcal{D}_i$ in $\mathbf{D}_\mathrm{s}$}{
			{$\mathcal{T}_\mathrm{o}^\mathrm{fine} \leftarrow \texttt{matching}(\mathcal{D}_i, \mathcal{T}_\mathrm{o}^\mathrm{fine}, \mathcal{T}_\mathrm{p})$ }
		}
		
		\BlankLine
		{$\mathbf{I}_\mathrm{s}, \mathbf{D}_\mathrm{s} = \{\mathrm{I}_k\}, \{\mathrm{D}_k\}$} \;
		{$\mathcal{T}_\mathrm{o}^\mathrm{coarse} \leftarrow \mathcal{T}_\mathrm{o}^\mathrm{fine}$} \;
		{$\mathcal{T}_\mathrm{o} \leftarrow \mathcal{T}_\mathrm{o}^\mathrm{fine}$} }
	\BlankLine
}
\BlankLine
{Return: $\mathcal{T}_\mathrm{o}$}\;
}
\end{algorithm}

\section{Benchmark Details}\label{sec:benchmak_details}
\subsection{Statistics}
\begin{figure}[t!]
	\begin{center}
		\includegraphics[width=0.475\textwidth]{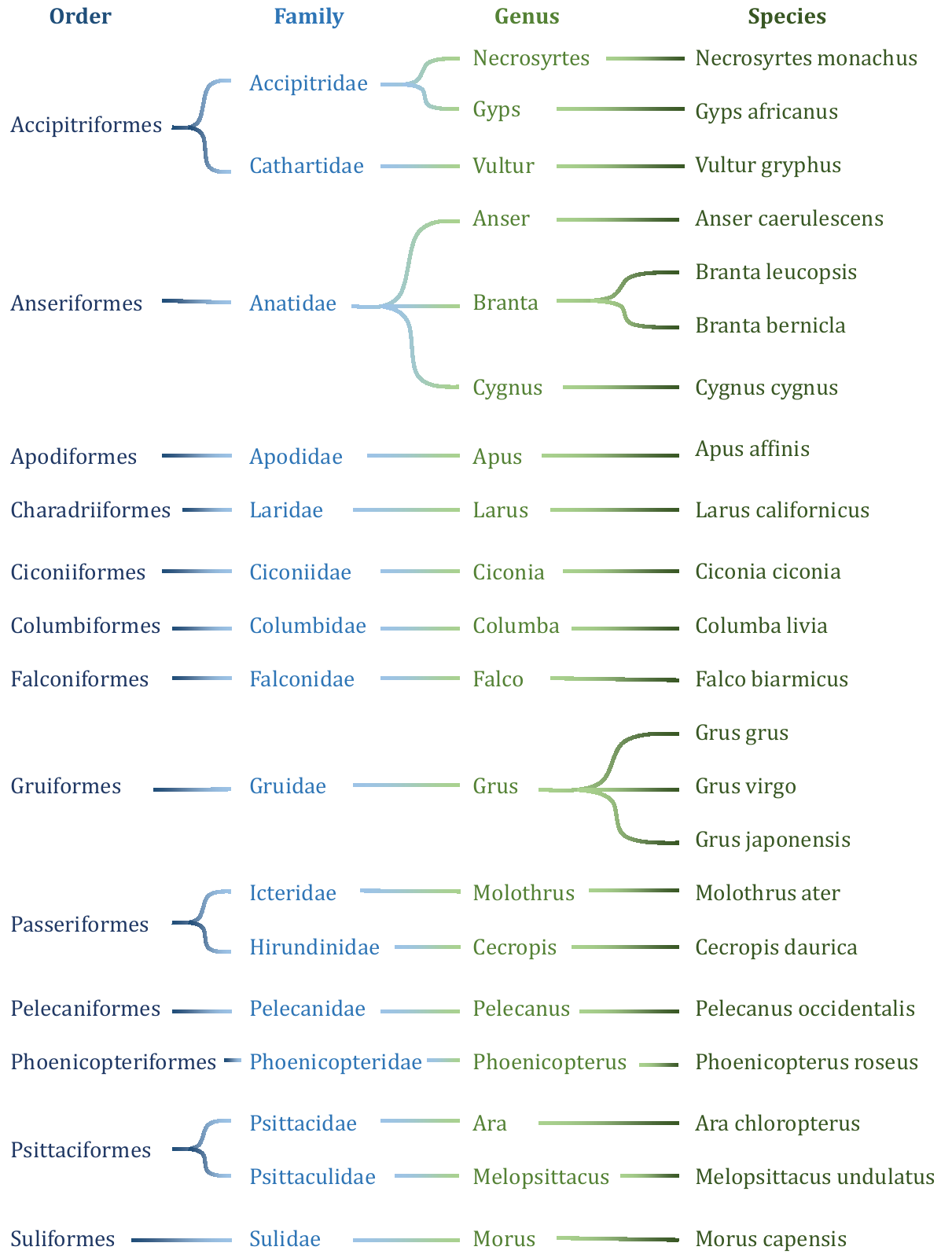}
	\end{center}
        \vspace{-10pt}
	\caption{Orders, families, genera, and species of objects in the BFT dataset, showcasing the diversity of the dataset.}
	\label{fig:order}
         \vspace{-10pt}
\end{figure}
\cref{fig:1}-c and \cref{fig:dataset} illustrate the geographical distribution of diverse scenes and environments in the dataset. The varied categories of objects in the BFT dataset also contribute to its diversity, as demonstrated in \cref{fig:order}, which includes 13 orders, 16 families, 19 genera, and 22 species.

\subsection{Dynamicity}
\begin{figure}[t!]
	\begin{center}                          
		\includegraphics[width=0.475\textwidth]{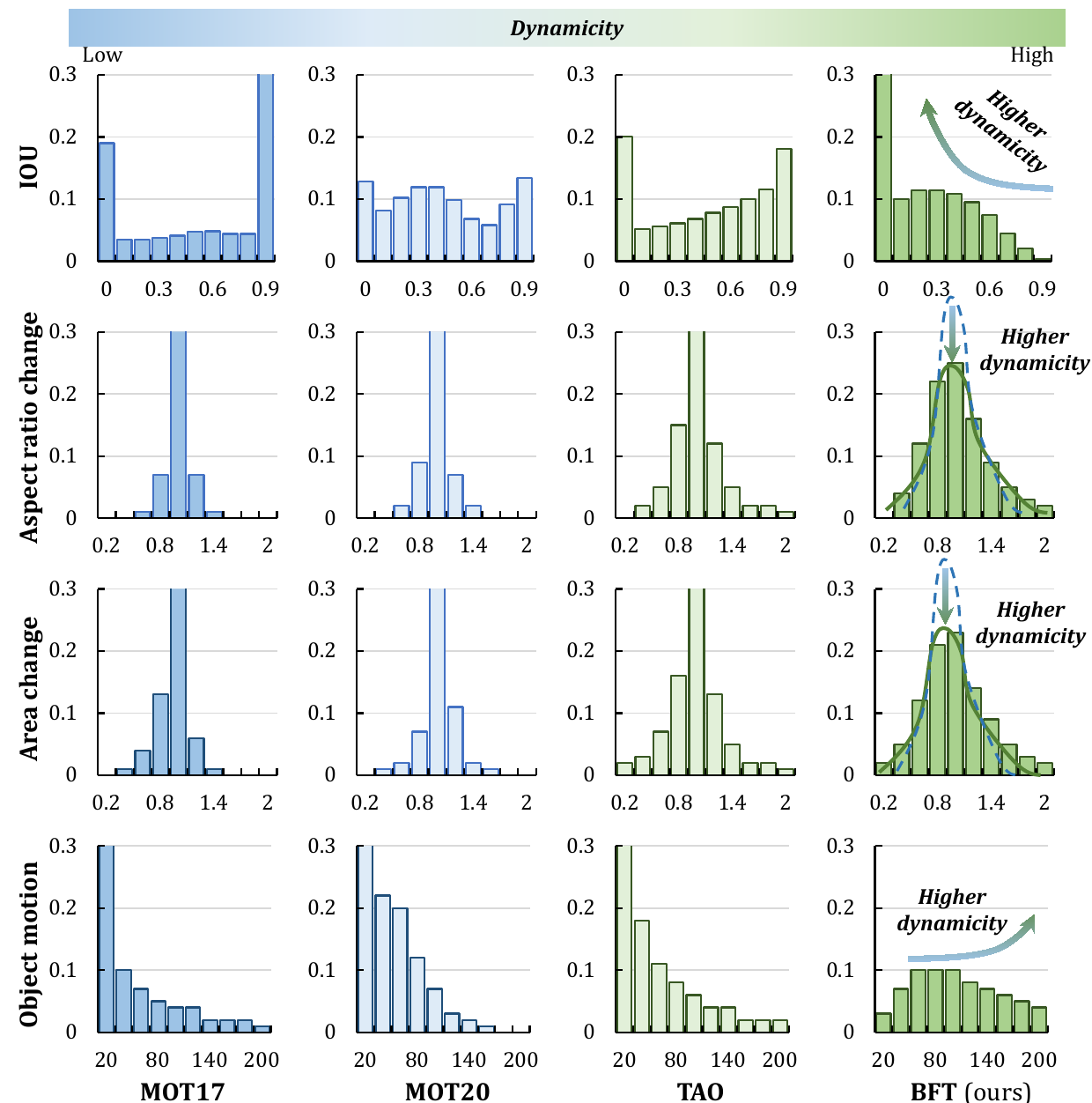}
	\end{center}
        \vspace{-10pt}
	\caption{Dynamicity comparison between BFT and other datasets on 4 attributes. BFT exhibits stronger dynamicity.}
	\label{fig:dynamicityv2}
         \vspace{-10pt}
\end{figure}

The dynamicity comparison between BFT and some open-world MOT datasets is shown in \cref{fig:dataset}-b,c, \cref{fig:dynamicityv2} further incorporates a comparison with closed-set datasets \cite{milan2016mot16, dendorfer2020mot20} and involves additional attributes to validate the dynamicity of open-world MOT compared to closed-set datasets, with particular emphasis on the pronounced dynamicity in BFT. All vertical axes in \cref{fig:dynamicityv2} represent frequency.

\noindent\textbf{IOU} The Intersection over Union (IOU) of the same object in adjacent frames can reflect the degree of geometric changes of the object, thus describing its dynamicity. Adjacent IOU is calculated by $\text{IOU}(\mathbf{b}^t, \mathbf{b}^{t-1}) = \frac{{|\mathbf{b}^t \cap \mathbf{b}^{t-1}|}}{{|\mathbf{b}^t \cup \mathbf{b}^{t-1}|}}$, where $\mathbf{b}$ is the bounding box of an object, and $t$ denotes frame index. In the first line of \cref{fig:dynamicityv2}, the interval of IOU is 0.1.
The lower IOU of BFT indicates its stronger dynamicity in the comparison.

\noindent\textbf{Aspect ratio change} Aspect ratio change (ARC) of the same object across adjacent frames reflects the object deformation in the horizontal and vertical directions, which is formulated as $\text{ARC}(\mathbf{b}^t, \mathbf{b}^{t-1}) = \frac{w^{t-1}/h^{t-1}}{w^t/h^t}$. $w$ and $h$ are the width and height of the bounding boxes, respectively. In the second line of \cref{fig:dynamicityv2}, the interval of ARC is 0.2.
The further the ARC deviates from 1, the more pronounced the dynamicity of the objects, indicating a greater significance of BFT's dynamicity.

\noindent\textbf{Area change} Area change (AC) of the object boxes between adjacent frames reflects the overall size variation of the object. AC is calculated as $\text{AC}(\mathbf{b}^t, \mathbf{b}^{t-1})=\frac{w^{t-1}h^{t-1}}{w^th^t}$.
In the third line of \cref{fig:dynamicityv2}, the interval of AC is 0.2.
BFT's dynamicity is more significant as AC deviates further from 1, resulting in a more pronounced dynamicity of objects.

\noindent\textbf{Object motion} The displacement of the center point of the object box between adjacent frames reflects the velocity of the object motion (OM). OM is formulated as $\text{OM}(\mathbf{b}^t, \mathbf{b}^{t-1}) = |x_c^{t} - x_c^{t-1}| + |y_c^{t} - y_c^{t-1}|$. $x_c$ and $y_c$ are the horizontal and vertical coordinates of the center of the bounding box, respectively. In the last line of \cref{fig:dynamicityv2}, the interval of OM is 20 pixels. In the BFT dataset, the overall OM is higher compared to other datasets, indicating a higher level of dynamicity within BFT.

\subsection{Visualization}
\begin{figure*}[t!]
	\begin{center}
		\includegraphics[width=0.95\textwidth]{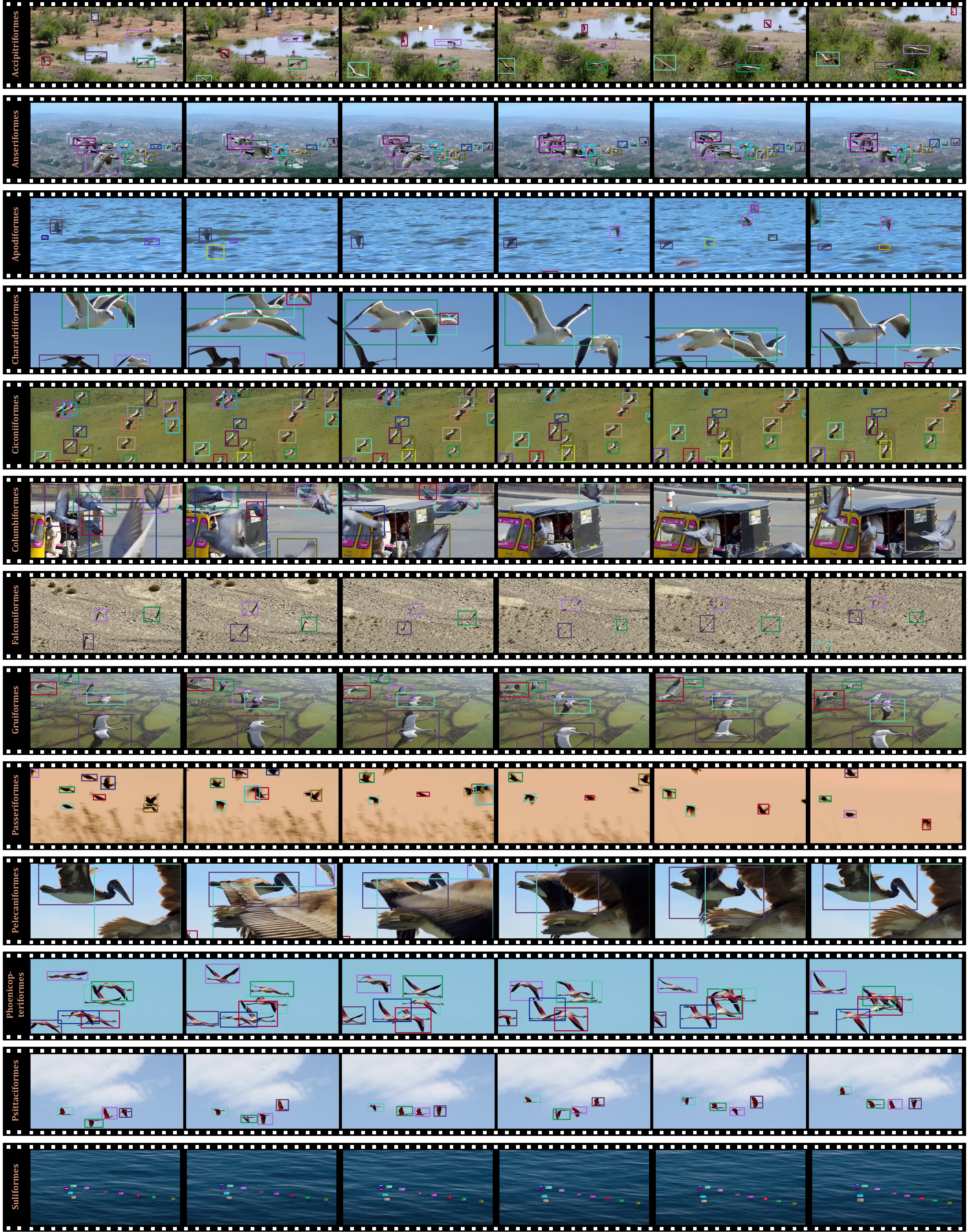}
	\end{center}
        \vspace{-10pt}
	\caption{Order-wise visualization of the BFT dataset, showcasing the diversity of the dataset and the dynamicity of tracking objects.}
	\label{fig:dataset_vis}
         \vspace{-10pt}
\end{figure*}

\begin{figure*}[t!]
	\begin{center}
		\includegraphics[width=0.95\textwidth]{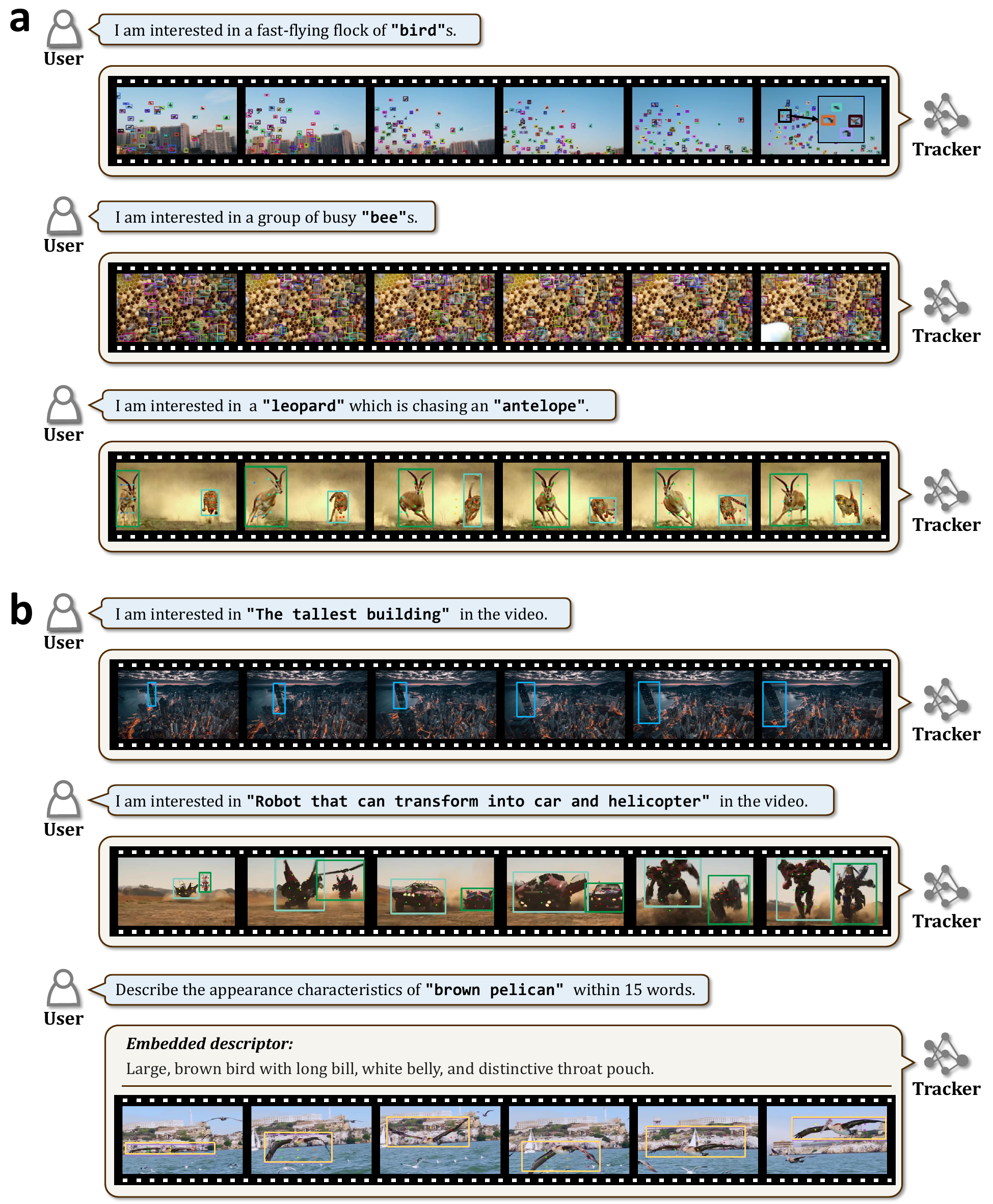}
	\end{center}
	\caption{Qualitative tracking results of NetTrack. \textbf{a} Scenarios involving highly dynamic objects and densely packed objects in open-world settings. \textbf{b} Understanding dynamic scenes and objects under referring expression comprehension conditions and domain-specific knowledge aided by embedded descriptors. }
	\label{fig:qualitative}
        \vspace{-10pt}
\end{figure*}

\begin{figure*}[t!]
	\begin{center}
		\includegraphics[width=0.95\textwidth]{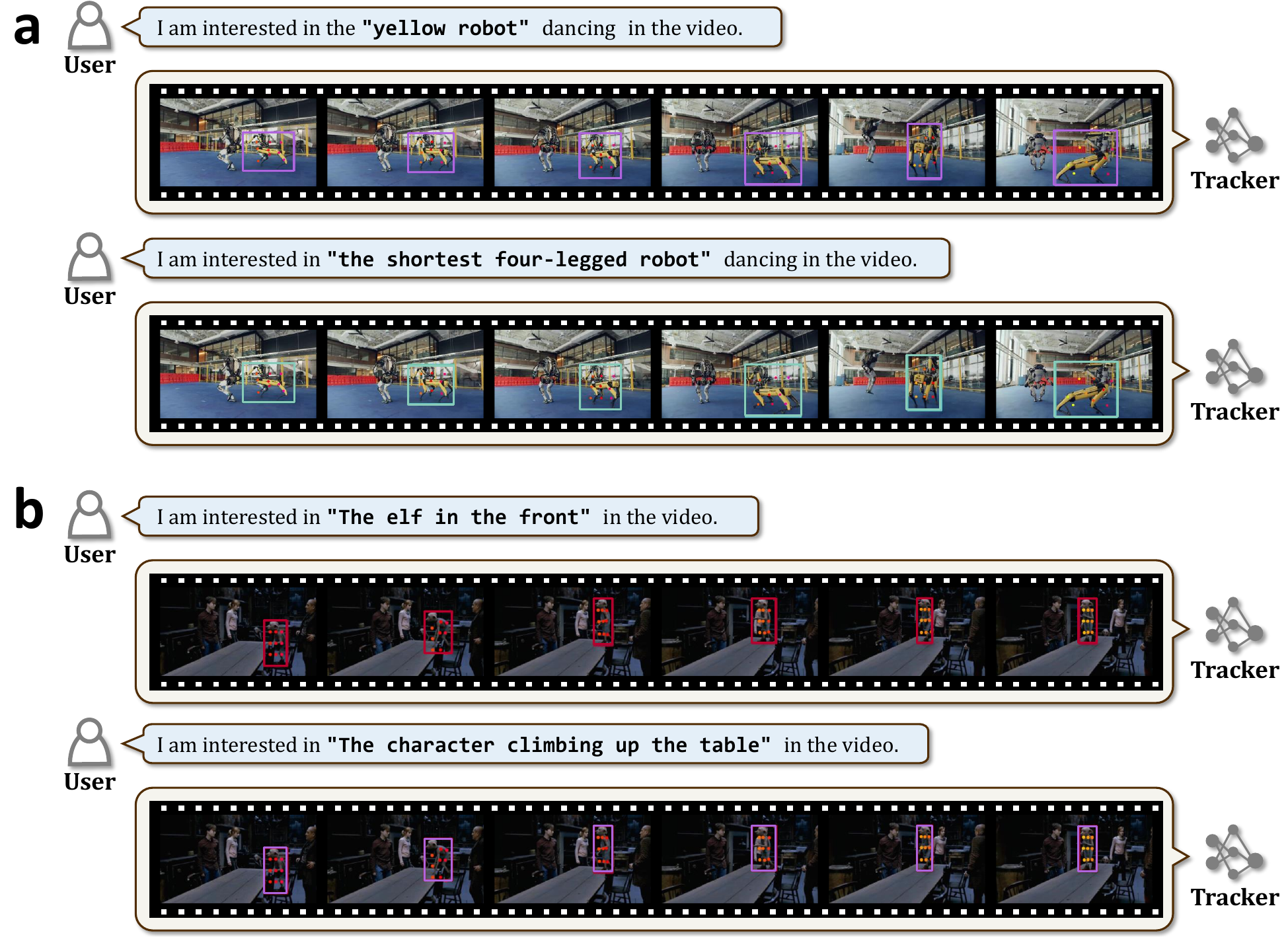}
	\end{center}
	\caption{Qualitative tracking results to validate the robustness of NetTrack under various prompts. \textbf{a} Describing the objects of interest from the perspectives of color and shape. \textbf{b} Describing the objects from the perspectives of appearance and behavior.}
	\label{fig:qualitative_prompt}
         \vspace{-10pt}
\end{figure*}

The visualization of representative videos contained in the BFT dataset is shown in \cref{fig:dataset_vis}, which covers all orders of birds in the dataset to demonstrate the diversity of data and the dynamicity of tracking objects.

\section{Experiment Details}\label{sec:exp_details}
\subsection{Dataset statistics}
\noindent\textbf{TAO} There are 295 overlapping classes between the 833 object classes in TAO~\cite{dave2020tao} validation set and LVIS~\cite{lvis}. Out of these, 35 classes are considered rare and are designated as \textit{novel} classes following OVTrack~\cite{ovtrack}. For evaluation purposes, there are a total of 109,963 annotations across 988 validation sequences, with 2,835 annotations belonging to \textit{novel} classes.

\noindent\textbf{TAO-OW} TAO-OW~\cite{tao-ow} validation set considers 52 COCO~\cite{coco} classes as \textit{known}, which consist of 87,358 distinct object tracks. In contrast, 209 classes that are not present in COCO are regarded as \textit{unknown}, comprising 20,522 distinct object tracks.

\noindent\textbf{AnimalTrack and GMOT-40} Following open-world settings~\cite{tao-ow}, the \textit{known} and \textit{unknown} classes of AnimalTrack~\cite{zhang2022animaltrack} test set and GMOT-40~\cite{bai2021gmot} are divided. In the AnimalTrack benchmark, the \textit{known} classes are \textit{horse} and \textit{zebra}, and the \textit{unknown} classes are \textit{chicken}, \textit{deer}, \textit{pig}, \textit{goose}, \textit{duck}, \textit{dolphin}, \textit{rabbit}, and \textit{penguin}. In the GMOT-40 benchmark, the \textit{known} classes are \textit{airplane}, \textit{bird}, \textit{boat}, \textit{sheep}, \textit{car}, and \textit{person}, and the unknown classes are \textit{helicopter}, \textit{billiard}, \textit{lantern}, \textit{tennis}, \textit{balloon}, \textit{fish}, \textit{bee}, \textit{duck}, \textit{penguin}, \textit{goat}, \textit{wolf}, and \textit{ant}.

\subsection{Metric details}
\noindent \textbf{OWTA} 
The \textit{open-world tracking accuracy} (OWTA)~\cite{tao-ow} consists of the \textit{detection recall} (DetRe) and \textit{association accuracy} (AssA). With a localization threshold $\alpha$, the OWTA metric is calculated as $\text{OWTA}_{\alpha}=\sqrt{\text{DetRe}_{{\alpha}} \cdot \text{AssA}_{\alpha}}$.
Specifically, DetRe is evaluated as $\text{DetRe}_{\alpha} = \frac{\text{TP}_\alpha}{\text{TP}_\alpha+\text{FN}_\alpha}$,
where the true positive (TP) and false negative (FP) are considered while false positives (FP) are not penalized. In AssA, the TP associations (TPA), FP associations (FPA), and FN associations (FNA) are incorporated into the calculation as $\text{AssA}_\alpha=\frac1{\text{TP}_\alpha}\sum_{c\in\text{TP}_\alpha}\mathcal{A}(c)$,
where $\mathcal{A}(c)$ is calucated as $\mathcal{A}(c)=\frac{\text{TPA}_\alpha(c)}{\text{TPA}_\alpha(c)+\text{FPA}_\alpha(c)+\text{FNA}_\alpha(c)}$.

\noindent \textbf{TETA}
The \textit{tracking-every-thing accuracy} (TETA)~\cite{tetr} consists of three components, the \textit{localization accuracy} (LocA), \textit{classification accuracy} (ClsA), and \textit{association accuracy} (AssocA). TETA is calculated as $\text{TETA}=\frac{\text{LocA}+\text{AssocA}+\text{ClsA}} 3$.
LocA is evaluated as $\text{LocA}=\frac{\text{TP}}{\text{TP}+\text{FP}+\text{FN}}$, 
and AssocA is derived in the same manner as AssA in OWTA.
For classification, ClsA is calculated as $\text{ClsA}=\frac{\text{TPC}}{\text{TPC}+\text{FPC}+\text{FNC}}$,
where TP classification (TPC), FP classification (FPC), and FN classification (FNC) are concerned. 

\noindent \textbf{HOTA, MOTA, and IDF1}
~~
These three metrics are for closed-set MOT and serve as a reference.
Higher Order Tracking Accuracy (HOTA)~\cite{HOTA} is calculated as $\text{HOTA}_\alpha=\sqrt{\text{DetA}_{{\alpha}} \cdot \text{AssA}_{\alpha}}$.
where the detection accuracy (DetA) is derived in the same manner as LocA in TETA.
Multiple object tracking accuracy (MOTA)~\cite{clear} measures the detection errors of FNs and FPs, as well as the association error of identification switch (IDSW), which is derived as $\text{MOTA}=1-\frac{\text{FN}+\text{FP}+\text{IDSW}}{\text{gtDet}}$,
where $\text{gtDet}$ refers to the number of groundtruth detections.
IDF1~\cite{idf1} is the ratio of correctly identified detections over the average number of ground-truth and computed detections, which balances identification precision and recall through harmonic mean and is calculated as $\text{IDF1}=\frac{2\text{IDTP}}{2\text{IDTP}+\text{IDFP}+\text{IDFN}}$,
where IDTP, IDFP, and IDFN refer to true positive, false positive, and false negative of identification, respectively.

\begin{figure}[t!]
	\begin{center}
		\includegraphics[width=0.475\textwidth]{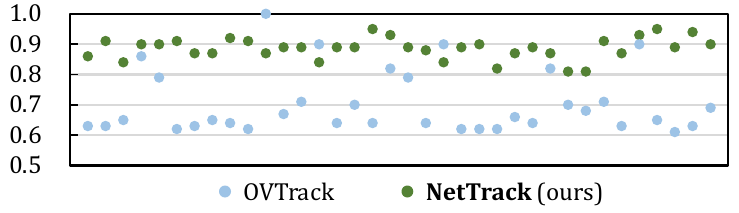}
	\end{center}
        \vspace{-10pt}
	\caption{LocA comparison between CLIP-based OVTrack and the proposed NetTrack on BFT benchmark. Each data point refers to a corresponding image sequence.}
	\label{fig:clip}
         \vspace{-10pt}
\end{figure}

\begin{figure}[t!]
	\begin{center}
		\includegraphics[width=0.475\textwidth]{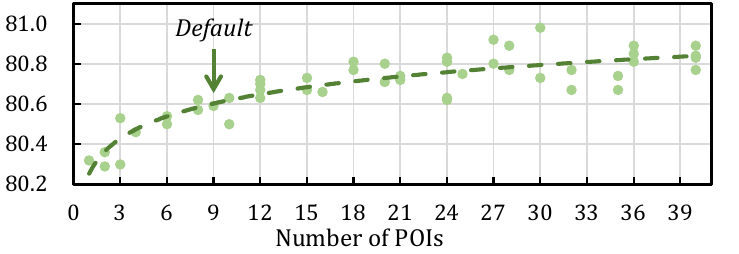}
	\end{center}
        \vspace{-10pt}
	\caption{D. Re. performance with different numbers of POIs. More POIs typically bring better performance and heavier computational burden. NetTrack aims to realize a trade-off between performance and efficiency.}
	\label{fig:grid}
         \vspace{-10pt}
\end{figure}

\begin{figure}[t!]
	\begin{center}
		\includegraphics[width=0.475\textwidth]{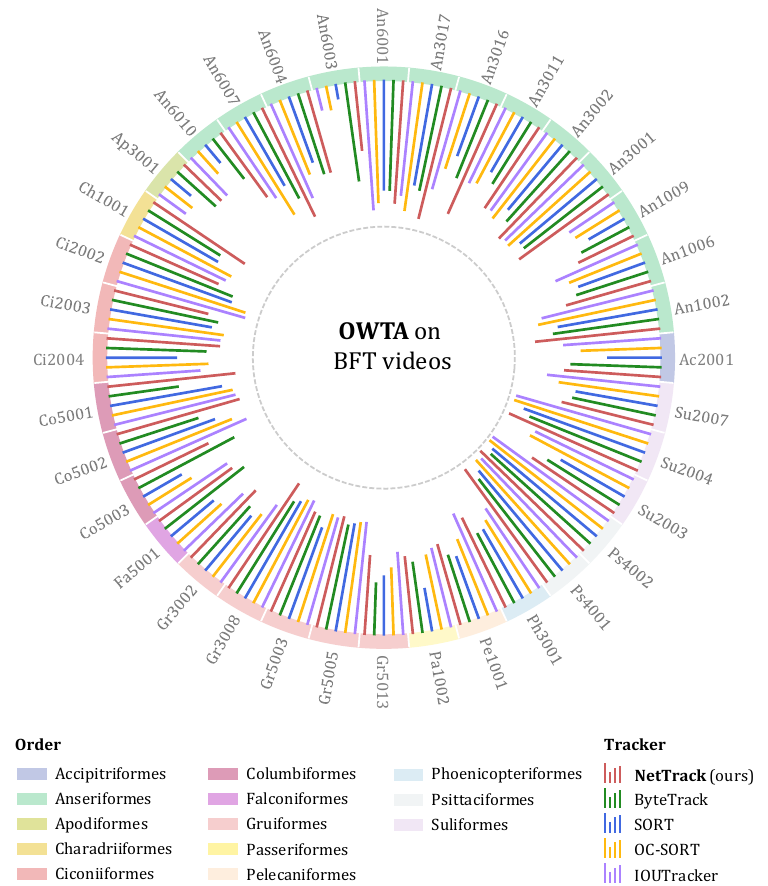}
	\end{center}
 \vspace{-10pt}
	\caption{Order-wise performance comparison. NetTrack exhibits stronger generalization ability on diverse categories.}
	\label{fig:order_wise}
     \vspace{-10pt}
\end{figure}

\subsection{Additional ablation studies}
\noindent\textbf{Comparison with CLIP-based pre-training} Compared to the previous coarse-grained CLIP~\cite{clip}-based method~\cite{ovtrack}, the proposed fine-grained object-text correspondence demonstrates better capability in localizing dynamic objects. As depicted in \cref{fig:clip}, each data point represents the performance of LocA in a sequence from the BFT dataset. Due to the excessive introduction of false positive samples by the compared method, the \textit{association accuracy} on dynamic targets is low (16.5) and thus not included in the comprehensive comparison. This comparison validates the applicability of the introduced fine-grained object-text correspondence to track highly dynamic objects.

\noindent\textbf{Number of POIs} Assigning excessive POIs to each potential object can lead to more robust performance, particularly in the ability to discover potential objects. However, it also results in a higher computational burden. \cref{fig:grid} illustrates the relationship between \textit{detection recall} and the number of POIs assigned to each object. Each specific POI count is decomposed into grids corresponding to different aspect ratios, \eg, 12 POIs can be decomposed into 3$\times$4, 4$\times$3, 2$\times$6, and 6$\times$2 grids, resulting in 4 data points. The aspect ratio of the grid is constrained within the range of [$\frac{1}{3}$, 3]. When the number of POIs exceeds 9, the performance improvement becomes marginal. Therefore, the default number of POIs is set to 9, corresponding to a 3$\times$3 grid.

\begin{figure}[t]
	\begin{center}
		\includegraphics[width=0.475\textwidth]{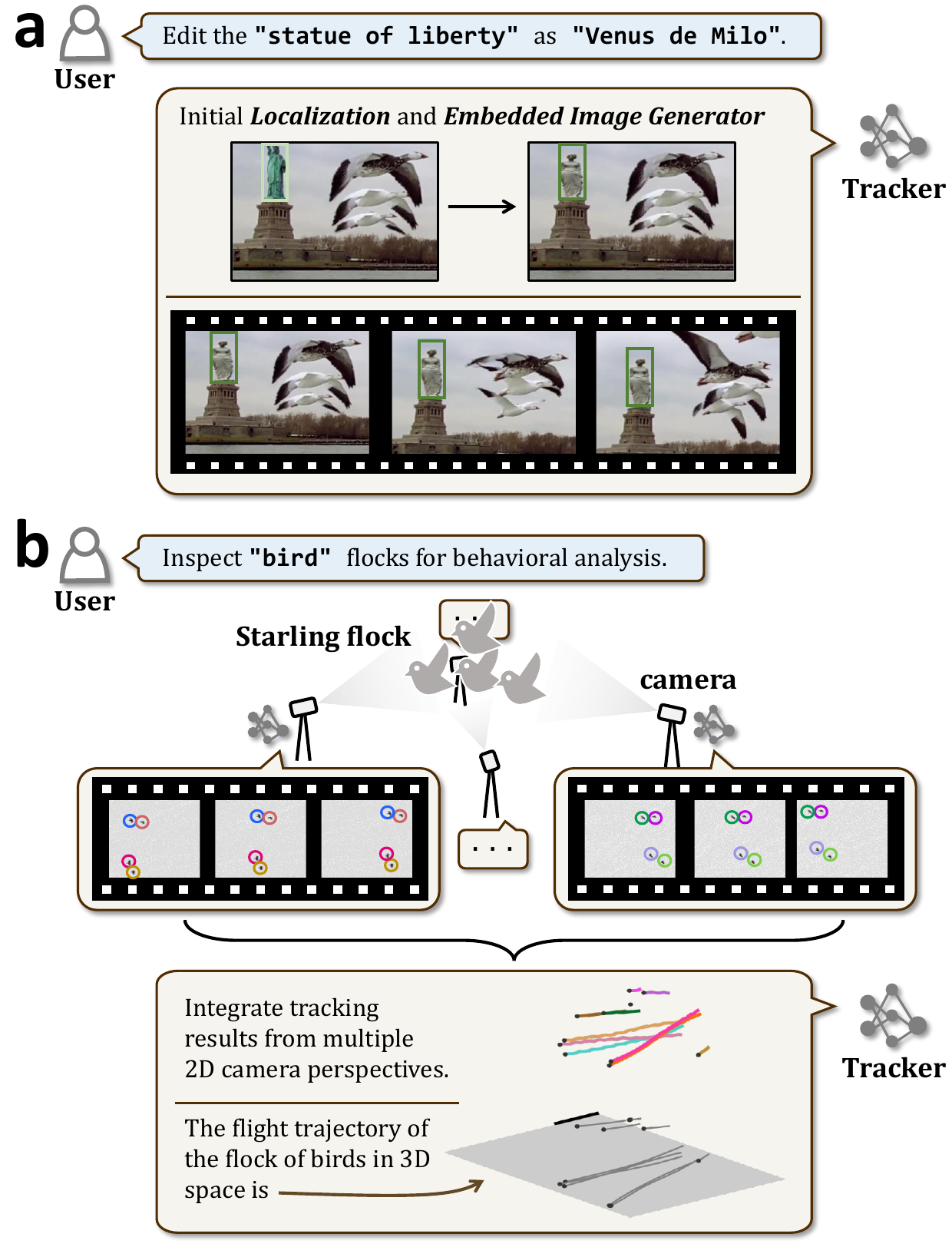}
	\end{center}
	\caption{Potential applications of NetTrack. \textbf{a} Video editing with embedded image generator, \eg, stable diffusion~\cite{stablediffusion}. \textbf{b} Ecological inspection for behavioral analysis, \eg, obtaining 3D trajectories of bird flocks through multi-view tracking results and 3D assignments with settings from~\cite{ling2018simultaneous}.}
	\label{fig:app}
        \vspace{-10pt}
\end{figure}

\noindent\textbf{Order-wise performance}
\cref{fig:order_wise} categorizes the test videos into the 13 orders included in the dataset and investigates the performance of the tracker on each order separately. NetTrack outperforms other SOTA trackers~\cite{sort,ocsort,bytetrack,ioutracker} on most orders, and the performance gap is not significant for different orders, which confirms its strong generalization ability.


\subsection{Qualitative results}
\label{sec:qualitative_supp}
\noindent\textbf{Tracking dynamic and dense objects} 
~~Tracking dynamic and numerous objects in an open-world environment poses significant challenges. \cref{fig:qualitative}-a illustrates NetTrack's performance in tracking dynamic and dense objects across three scenarios, including fast-moving and numerous flocks of birds and bees, as well as highly deformable objects like leopards and antelope. Despite these challenges, NetTrack demonstrates excellent robustness.

\noindent\textbf{Referring expression comprehension} The understanding of dynamic scenes in open-world environments is crucial for the practical application of trackers in MOT. \cref{fig:qualitative}-b illustrates three scenarios: continuously understanding and tracking the tallest buildings, tracking constantly deforming Transformers, and learning professional knowledge through an embedded descriptor~\cite{gpt}. NetTrack demonstrates its ability to comprehend dynamic scenes and its potential value in practical applications.

\noindent\textbf{Robustness to various prompts} In practical applications, the user's prompt input may be biased due to different focuses, making it important for the tracker to have robust performance with diverse prompts. \cref{fig:qualitative_prompt} demonstrates scenarios where the object of interest remains the same, but two different prompts are given. In scenario \textbf{a}, prompts focus on the color and height of the robot, respectively, while in scenario \textbf{b}, prompts focus on the category of the character and the ongoing action. Faced with diverse prompts, NetTrack is able to maintain robust performance, confirming its potential in practical applications.


\section{Applications}\label{sec:app}
In addition to the use of embedded descriptors (\eg, large language models~\cite{gpt,gpt4}) to understand domain-specific knowledge as shown in \cref{fig:qualitative}-b, \cref{fig:app} also demonstrates two other potential applications of NetTrack. In \cref{fig:app}-a, an embedded image generator performs inpainting on the objects of interest for video editing. The tracker first locates the objects of interest and then uses the image generator (stable diffusion~\cite{stablediffusion} in this example) to inpaint on the area of interest, achieving the desired video editing effect. Furthermore, in \cref{fig:app}-b, due to NetTrack's use of fine-grained representations of objects, even small and dynamic objects can be tracked by point tracking, enabling the acquisition of three-dimensional (3D) trajectories for ecological inspection, such as bird flight trajectories through multi-view tracking results and 3D assignment. The data and experimental settings for ecological inspection in \cref{fig:app} are sourced from~\cite{ling2018simultaneous}. It is worth mentioning that these are just brief descriptions of some potential applications of NetTrack. Due to its strong generalization ability, better integration with foundation models will lead to even broader and more practical application value.

\end{document}